\def\year{2020}\relax
\documentclass[letterpaper]{article} 
\usepackage{aaai20}  
\usepackage{times}  
\usepackage{helvet} 
\usepackage{courier}  
\usepackage[hyphens]{url}  
\usepackage{graphicx} 
\usepackage{graphicx}  
\usepackage[table]{xcolor}

\newcommand{\R}{\mathbb{R}}

\newcommand{\mylist}[1]{\langle{}#1\rangle{}}
\newcommand{\abs}[1]{\left|#1\right|}
\newcommand{\N}{\mathbb{N}}

\newcommand{\true}{\mathtt{t \hspace{-.5mm}t}}
\newcommand{\false}{\mathtt{f\hspace{-.5mm}f}}
\newcommand{\Autom}{A}

\newcommand{\regr}[1]{\ensuremath{\mathbf{RGR}(#1)}}
\newcommand{\search}[1]{\ensuremath{\mathbf{BFS}(#1)}}

\newcommand{\citet}[1]{\citeauthor{#1}~\shortcite{#1}}

\newcommand{\transpose}[1]{#1^{\top}}

\newcommand{\ConfMapSymbol}[1]{\delta_{#1}}
\newcommand{\WeightMapSymbol}[1]{f_{#1}}
\newcommand{\ConfMap}[2]{\ConfMapSymbol{#1}(#2)}
\newcommand{\WeightMap}[2]{\WeightMapSymbol{#1}(#2)}

\newcommand{\WFAstate}{\ensuremath{q}}
\newcommand{\WFAState}{\ensuremath{Q_{\Autom}}}
\newcommand{\WFAInitState}{\ensuremath{\alpha_{\Autom}}}
\newcommand{\WFAAccState}{\ensuremath{\beta_{\Autom}}}
\newcommand{\WFATransition}[1][\sigma]{A_{#1}}
\newcommand{\WFAAllTransition}[1]{\ensuremath{(\WFATransition[\sigma])_{\sigma \in \Sigma}}}
\newcommand{\WFAconf}{x}
\newcommand{\WFAConf}{\R^{\WFAState}}
\newcommand{\WFAConfMapSymbol}{\delta_{\Autom}}
\newcommand{\WFAWeightMapSymbol}{f_{\Autom}}

\newcommand{\DFAstate}{\ensuremath{q}}
\newcommand{\DFAState}{\ensuremath{Q_{\Autom}}}
\newcommand{\DFAInitState}{\ensuremath{\alpha_{\Autom}}}

\newcommand{\DFATransition}[1][\sigma]{A_{#1}}
\newcommand{\DFAAllTransition}[1]{\ensuremath{(\DFATransition[\sigma])_{\sigma \in \Sigma}}}

\newcommand{\DFAConf}{\{\true,\false\}^{\DFAState}}
\newcommand{\DFAConfMapSymbol}{\WFAConfMapSymbol}
\newcommand{\DFAWeightMapSymbol}{\WFAWeightMapSymbol}
\newcommand{\DFAConfMap}[1]{\DFAConfMapSymbol(#1)}
\newcommand{\DFAWeightMap}[1]{\DFAWeightMapSymbol(#1)}

\newcommand{\lstar}{$\mathrm{L}^{*}$}

\usepackage{algorithm}
\usepackage[noend]{algpseudocode}
\renewcommand{\Return}[1]{\State \textbf{return} #1}
\newcommand{\Break}{\State \textbf{break}}

\algrenewcommand\algorithmicindent{1em}%

\usepackage{multirow}
\usepackage{braket}
\usepackage{amsmath}
\usepackage{amsfonts}
\usepackage{amsthm}
\usepackage[capitalize]{cleveref}
\usepackage{booktabs}       
\usepackage{subcaption}
\usepackage{wrapfig}
\usepackage{paralist} 

\newtheorem{theorem}{Theorem}

\theoremstyle{definition}

\newtheorem{example}[theorem]{Example}
\newtheorem{definition}[theorem]{Definition}

\ifdefined \VersionCameraReady
\newcommand{\CameraReadyVersion}[1]{#1}
\newcommand{\ArxivVersion}[1]{}
\else
\newcommand{\CameraReadyVersion}[1]{}
\newcommand{\ArxivVersion}[1]{#1}
\fi
\newif\ifignore 
\ignorefalse
\newcommand{\auxproof}[1]{
  \ifignore\mbox{}\newline
  \textbf{BEGIN: AUX-PROOF} \dotfill\newline
  {#1}\mbox{}\newline
  \textbf{END: AUX-PROOF}\dotfill\newline
  \fi}

\ifignore
	\usepackage[colorinlistoftodos,textsize=footnotesize]{todonotes}
\else
	\usepackage[disable]{todonotes}
\fi
\newcommand{\gennote}[3]{\todo[inline,linecolor=#2,backgroundcolor=#2!25,bordercolor=#2]{#3: #1}}

\newcommand{\masaki}[1]{\gennote{#1}{orange}{Masaki}}

\newcommand{\taro}[1]{\gennote{#1}{green}{Taro}}

\newcommand{\TODO}[1]{\gennote{#1}{red}{TODO}}
\newcommand{\TS}[1]{\textcolor{blue}{TS: #1}}

\title{Weighted Automata Extraction from Recurrent Neural Networks \\ via Regression on State Spaces}
\author{Takamasa Okudono, Masaki Waga, Taro Sekiyama, Ichiro Hasuo\\
National Institute of Informatics \&  The Graduate University for Advanced Studies\\
2-1-2 Hitotsubashi, Chiyoda-ku, Tokyo 101-8430\\
\{tokudono,mwaga,sekiyama,hasuo\}@nii.ac.jp
}
\usetikzlibrary{arrows,automata,positioning}
 
\begin{document}

\maketitle

\begin{abstract}
    We present a method to extract a weighted finite automaton (WFA) from a
    recurrent neural network (RNN). Our method is based on the WFA learning
    algorithm by Balle and Mohri, which is in turn an extension of Angluin's
    classic \lstar algorithm. Our technical novelty is in the use of
    \emph{regression} methods for the so-called equivalence queries, thus
    exploiting the internal state space of an RNN to prioritize counterexample
    candidates.  This way we achieve a quantitative/weighted extension of the recent work
    by Weiss, Goldberg and Yahav that extracts DFAs.  
    We experimentally evaluate the accuracy, expressivity and efficiency of the extracted WFAs.
\end{abstract}

\masaki{I changed the comments to ``inline'' because the margin is too small to use ``not-inline'' mode. }

\section{Introduction}
\label{sec:intro}
%
\subsubsection{Background}
\emph{Deep neural networks (DNNs)} 
have been successfully applied to domains such as text, speech,
and image processing.  \emph{Recurrent neural networks
(RNNs)}~\cite{DBLP:journals/corr/ChungGCB14,DBLP:journals/neco/HochreiterS97}
is a class of DNNs equipped with the capability of processing sequential data of
variable length.  The great success of RNNs has been seen in, 
e.g., 
machine translation~\cite{DBLP:conf/nips/SutskeverVL14}, 
speech recognition~\cite{DBLP:conf/icassp/ZweigYDS17}, and anomaly
detection~\cite{DBLP:journals/access/LvWYL18,DBLP:journals/mta/XiaoZMHS19}.

While it has been experimentally shown that RNNs are a powerful tool to process,
predict, and model sequential data, there are known drawbacks in RNNs such as interpretability and costly inference.
A research line that attacks this challenge is \emph{automata extraction}~\cite{OmlinG96,DBLP:conf/icml/WeissGY18}. Focusing on RNNs' use as \emph{acceptors} (i.e., receiving an input sequence and producing a single Boolean output), these works extract a \emph{finite-state automaton} from an RNN as a succinct and interpretable surrogate.
Automata extraction exposes internal transition between the states of an RNN in the form of an
automaton, which is then amenable to  algorithmic analyses such as reachability and model checking~\cite{BaierK08}.
 Automata extraction can also be seen as \emph{model compression}: finite-state automata are usually more compact, and cheaper to run, than neural networks. 

\subsubsection{Extracting WFAs from RNNs}
Most of the existing automata extraction techniques  target Boolean-output RNNs, which however excludes many applications. 
In sentiment analysis, it is  desired to know
the quantitative strength of sentiment, besides its (Boolean) existence~\cite{DBLP:series/sbcs/Chaudhuri19}.
%
%
%
 RNNs with real values as their output are also useful  in classification tasks.  For
example, predicting class probabilities is a key in some approaches to
semi-supervised learning~\cite{Yarowsky95} and ensemble~\cite{Bramer2013}.

This motivates extraction of \emph{quantitative} finite-state machines  as abstraction of RNNs. We find the formalism of \emph{weighted finite automata} (WFAs) suited for this purpose. A WFA is a finite-state machine---much like a deterministic finite automaton (DFA)---but its transitions as well as acceptance values are real numbers (instead of Booleans). 

\subsubsection{Contribution: Regression-Based WFA Extraction from RNNs}
Our main contribution is a procedure that takes a (real-output) RNN $R$, and
returns a WFA $A_{R}$ that abstracts $R$.  The procedure is based on the WFA
learning algorithm in~\cite{DBLP:conf/cai/BalleM15}, that is in turn based on
the famous \emph{\lstar~algorithm} for learning
DFAs~\cite{DBLP:journals/iandc/Angluin87}.
These algorithms learn automata by a series of so-called \emph{membership queries} and \emph{equivalence queries}.
In our procedure, a membership query is implemented by an inference of the given RNN $R$.
We iterate membership queries and use their results to construct a WFA $A$ and to
grow it. 

The role of equivalence queries is to say when to stop this iteration:
it asks if $A$ and $R$ are ``equivalent,'' that is, if the WFA $A$ obtained so
far is comprehensive enough to cover all the possible behaviors of $R$. 
This is not possible in general---RNNs are more expressive than WFAs---therefore we inevitably resort to an approximate method.
Our
technical novelty lies in the method for answering equivalence queries;
notably it uses a \emph{regression} method---e.g., the 
Gaussian process regression (GPR) and the kernel ridge regression (KRR)---for
abstraction of the state space of $R$.

We conducted experiments to evaluate the effectiveness of our approach. 
In particular, we are concerned with the following questions:
\begin{inparaenum}[1)]
 \item \label{Q:acc} how similar the behavior of the extracted WFA $A_{R}$, and that of the original RNN $R$, are;
 \item \label{Q:exp} how applicable our method is to an RNN that is a more expressive model than WFAs; and
 \item \label{Q:eff} how efficient the inference of the WFA $A_{R}$ is, when compared with the inference of $R$.
\end{inparaenum}
The experiments we designed for the questions~\ref{Q:acc}) and~\ref{Q:eff}) are with
RNNs trained using randomly generated WFAs.  The results show, on
the question~\ref{Q:acc}), that the WFAs extracted by our method approximate the
original RNNs accurately. This is especially so when compared with a baseline algorithm (a straightforward adaptation of \citet{DBLP:conf/cai/BalleM15}). On the question~\ref{Q:eff}), the inference of
the extracted WFAs are about 1300 times faster than that of the original RNNs.
On the question~\ref{Q:exp}), we devised an RNN that models a weighted variant of a (non-regular) language of balanced parentheses.
Although this weighted language is
beyond  WFAs' expressivity, we found that our method extracts
 WFAs that successfully approximate the RNN up-to a certain depth bound.

The paper is organized as follows. Angluin's \lstar~algorithm and its weighted adaptation are recalled  in~\S\ref{sec:prelim}. Our WFA extraction procedure is described in~\S\ref{sec:ourAlgorithm} focusing on our novelty, namely the regression-based procedure for answering equivalence queries.   Comparison with the DFA extraction by \citet{DBLP:conf/icml/WeissGY18} is given there, too. In~\S{}\ref{sec:implExpr} we discuss our experiment results.

\subsubsection{Potential Applications}
A major potential application of WFA extraction is to analyze an RNN $R$ via its interpretable surrogate $A_{R}$. The theory of WFAs offers a number of analysis methods, such as  \emph{bisimulation metric}~\cite{BalleGP17}---a distance notion between WFAs---that will allow us to tell how apart two RNNs are.

Another major potential application 
is as ``a poor
man's RNN $R$.'' While RNN inference is recognized to be rather expensive (especially for edge devices), simpler models by WFAs should be cheaper to run. 
Indeed, our
experimental results show (\S{}\ref{sec:implExpr}) that WFA inference is 
 about 1300
times faster than inference of original  RNNs.  


%
Since WFAs are defined over a finite alphabet, we  restrict to RNNs $R$ that take sequences over a \emph{finite} alphabet. This restriction should not severely limit the applicability of our method. Indeed, such RNNs (over a finite alphabet) have successfully  applications in many domains, including
intrusion prediction~\cite{DBLP:journals/access/LvWYL18}, 
malware detection~\cite{DBLP:journals/mta/XiaoZMHS19}, and 
DNA-protein binding prediction~\cite{shen2018recurrent}. Moreover, even if  inputs are    real numbers, \emph{quantization}   is commonly employed without losing a lot of precision. See, e.g., recent~\cite{GuptaAGN15}.

\subsubsection{Related Work}
The relationship between RNNs and  automata
 has been studied in both
non-quantitative~\cite{OmlinG96,DBLP:conf/acl/WeissGY18,DBLP:conf/icml/WeissGY18,DBLP:conf/iclr/MichalenkoSVBCP19} and
quantitative~\cite{DBLP:conf/icgi/AyacheEG18,DBLP:conf/aistats/RabusseauLP19} settings.
Some of these works feature automata extraction from RNNs; we shall now discuss recent ones among them.

The work by \citet{DBLP:conf/icml/WeissGY18} is a pioneer in automata extraction from RNNs. They extract DFAs from RNNs, using a variation of \lstar algorithm, much like this work.
We provide a systematic comparison in~\S{}\ref{subsec:comparisonWithWeissEtAl}, identifying some notable similarities and differences.


\citet{DBLP:conf/icgi/AyacheEG18} extract a WFA from a black-box sequence acceptor whose example is an RNN.
Their method does not use equivalence queries; in contrast,
we exploit the internal state space of an RNN to approximately answer  equivalence queries.

DeepStellar~\cite{DBLP:conf/sigsoft/DuXLM0Z19} extracts Markov chains from RNNs, and uses them for coverage-based  testing and adversarial sample detection. Their extraction,  differently from our \lstar-like method, uses profiling from the training data and discrete abstraction of the state space. 

\citet{DBLP:conf/icml/WangN19} propose a new RNN architecture that makes it easier to extract a DFA from a trained model.  To apply their method, one has to modify the structure of an RNN before training, while our method does not need any special structure to RNNs and can be applied to already trained RNNs.

\citet{schwartz-etal-2018-bridging} introduce a neural network architecture that can represent (restricted forms of) CNNs and RNNs.  WFAs could also be expressed by their architecture, but extraction of automata is out of their interest.


%
A major approach to optimizing neural networks is by compression: 
\emph{model pruning}~\cite{HanPTD15}, \emph{quantization}~\cite{GuptaAGN15}, and
\emph{distillation}~\cite{BucilaCN06}. 
Combination and  comparison with these techniques is interesting future work.

\section{Preliminaries}
\label{sec:prelim}
We fix a finite alphabet $\Sigma$. The set of (finite-length) words over $\Sigma$ is  $\Sigma^{*}$. The \emph{empty word} (of length $0$) is denoted by $\varepsilon$. The length of a word $w\in \Sigma^*$ is denoted by $|w|$. 

\masaki{In case we omit the definition of DFAs, we can rename the paragraph name to e.g., ``Weighted Finite Automata''.}
We recall basic notions on WFAs. See~\cite{DrosteKV09} for details.

\begin{definition}[WFA]
 \label{def:WFA}
 A \emph{weighted finite automaton} (WFA) over $\Sigma$  is
 a quadruple 
\begin{math}
 \Autom=\bigl(\WFAState, \WFAInitState, \WFAAccState, \WFAAllTransition\bigr)
\end{math}.
Here
 $\WFAState$ is a finite set of \emph{states}; 
 $\WFAInitState, \WFAAccState$ are row vectors of size $|\WFAState|$ called the \emph{initial} and \emph{final} vectors; and
 $\WFATransition$ is a \emph{transition} matrix of $\sigma$, given for each $\sigma\in\Sigma$.
 For each $\sigma\in\Sigma$, $\WFATransition$ is a matrix of size $|\WFAState|\times|\WFAState|$.
\end{definition}

\begin{definition}[configuration of a WFA]
 \label{def:WFAConfig}
 Let $\Autom$ be the WFA in Def.~\ref{def:WFA}.
 A \emph{configuration} of $\Autom$ is a row vector $\WFAconf \in \WFAConf$.
For a word $w=\sigma_{1}\sigma_{2}\dotsc \sigma_{n}\in \Sigma^{*}$ (where $\sigma_{i}\in \Sigma$), the \emph{configuration} of $\Autom$ at $w$ is defined by 
\begin{math}
\textstyle \ConfMap{\Autom}{w} = \transpose{\WFAInitState} \cdot \bigl(\prod_{i=1}^{n} \WFATransition[\sigma_i]\bigr)
\end{math}.
\end{definition}
Obviously $\ConfMap{\Autom}{w} \in \WFAConf$ is a row vector of size $|\WFAState|$; it records the  weight at each state $\WFAstate \in \WFAState$ after reading $w$.

\begin{definition}
 [weight $\WeightMap{\Autom}{w}$ of a word in a WFA]
 \label{def:WFAWeightOfAWord}
 Let a WFA $\Autom$ and a word $w=\sigma_{1}\dotsc \sigma_{n}$ be  as in Def.~\ref{def:WFAConfig}. 
 The \emph{weight} of $w$ in $\Autom$ is given by
\begin{math}
\textstyle \WeightMap{\Autom}{w} = \transpose{\WFAInitState} \cdot\bigl(\prod_{i=1}^n \WFATransition[\sigma_i] \bigr)\cdot \WFAAccState
\end{math},
multiplying the final vector to the configuration at $w$. 
\end{definition}

\begin{example}
 \label{example:WFA}
 Let
 $\Sigma = \{a,b\}$,
 $\WFAState = \{\WFAstate_1, \WFAstate_2, \WFAstate_3\}$,
 $\WFAInitState = \transpose{(1\  2\  3)}$,
 $\WFAAccState = \transpose{(0\ {-1}\  1)}$, 
 $\WFATransition[a] = \left(
 \begin{array}[c]{c c c}
  1& 2& -1\\
  3& 0& 0\\
  0& 4& 0\\
 \end{array}\right)$, and
 $\WFATransition[b] = \left(
 \begin{array}[c]{c c c}
  -1& 1&0\\
  0& 3& 0\\
  -2& 4& 0\\
 \end{array}\right)$.
 For
 the WFA $\Autom = (\WFAState,\WFAInitState,\WFAAccState,(\WFATransition[\sigma])_{\sigma\in\Sigma})$ over $\Sigma$, and 
 $w = ba$, the configuration $\ConfMap{\Autom}{w}$ and the weight $\WeightMap{\Autom}{w}$ are as follows.
 \begin{align*}
  \ConfMap{\Autom}{w}
  &= \transpose{\WFAInitState} \WFATransition[b] \WFATransition[a] \\
  &= 
  \begin{pmatrix}
    1& 2& 3\\
  \end{pmatrix}
  \begin{pmatrix}
    -1& 1&0\\
  0& 3& 0\\
  -2& 4& 0
  \end{pmatrix}
  \begin{pmatrix}
    1& 2& -1\\
    3& 0& 0\\
    0& 4& 0
  \end{pmatrix}\\
  &= 
  \begin{pmatrix}
    50& -14& 7\\
  \end{pmatrix}
 \end{align*}
 \begin{align*}
  &\WeightMap{\Autom}{w}
  = \transpose{\WFAInitState} \WFATransition[b] \WFATransition[a] \WFAAccState\\
  &= 
  \begin{pmatrix}
    1& 2& 3\\
  \end{pmatrix}
  \begin{pmatrix}
    -1& 1&0\\
    0& 3& 0\\
    -2& 4& 0
  \end{pmatrix}
  \begin{pmatrix}
    1& 2& -1\\
    3& 0& 0\\
    0& 4& 0
  \end{pmatrix}
  \begin{pmatrix}
    0\\ -1\\ 1
  \end{pmatrix}\\
  &=21
 \end{align*}
 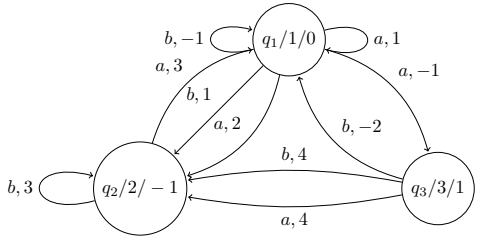
\begin{figure}[tbp]
  \centering
 \scalebox{0.7}{ 
 \begin{tikzpicture}[shorten >=1pt,node distance=4.0cm,on grid,auto] 
 \node[state] (q_0) {$\WFAstate_1/1/0$};
 \node[state] (q_1)[below left=of q_0] {$\WFAstate_2/2/-1$};
 \node[state] (q_2)[below right=of q_0] {$\WFAstate_3/3/1$};

 \path[->] 
  (q_0) edge [loop right] node {$a, 1$} (q_0)
  (q_0) edge [loop left] node {$b, -1$} (q_0)
  (q_0) edge [bend left] node[above left] {$a, 2$} (q_1)
  (q_0) edge node[above left] {$b, 1$} (q_1)
  (q_0) edge [bend left] node {$a, -1$} (q_2)
  (q_1) edge [bend left] node {$a, 3$} (q_0)
  (q_1) edge [loop left] node {$b, 3$} (q_1)
  (q_2) edge [bend left] node[above right] {$b, -2$} (q_0)
  (q_2) edge [bend left=10] node {$a, 4$} (q_1)
  (q_2) edge [bend right=10] node[above] {$b, 4$} (q_1)
  ;
 \end{tikzpicture}}
  \caption{An illustration of WFA $\Autom$ in Example~\ref{example:WFA}.
  In a state label ``$q/m/n$'', $q$ is a state name and $m$ and $n$ are the initial and final values at $q$, respectively.
  In the label ``$\sigma, p$'' of the transition from $q_i$ to $q_j$, $p$ is $\WFATransition[\sigma][i,j]$, where $\sigma \in \Sigma$ and $\WFATransition[\sigma][i,j]$ is the entry of $\WFATransition[\sigma]$ at row $i$ and column $j$.
  }
  \label{fig:WFA}
 \end{figure}
 Fig.~\ref{fig:WFA} illustrates the WFA $\Autom = (\WFAState,\WFAInitState,\WFAAccState,(\WFATransition[\sigma])_{\sigma\in\Sigma})$, where the transitions with weight $0$ are omitted.
\end{example}


\masaki{
If we need more space, we can move Definition~\ref{def:DFA} to the appendix. 
In that case, we should also rename the paragraph name.}
\begin{definition}[DFA]
 \label{def:DFA}
 A \emph{DFA} is defined much like in Def.~\ref{def:WFA}, except that
\begin{inparaenum}[1)]
 \item the entries of matrices  are $\true$ and $\false$; 
 \item we replace the use of $+,\times$  with $\lor,\land$, respectively; and
 \item we impose \emph{determinacy},  that exactly one entry is $\true$ in each row of $\DFATransition$, and that only one entry is $\true$ in the initial vector $\DFAInitState$.
\end{inparaenum}

The definitions of $\ConfMapSymbol{\Autom}$ and $\WeightMapSymbol{\Autom}$ in Def.~\ref{def:WFAConfig}--\ref{def:WFAWeightOfAWord}  adapt to DFAs.
For $w\in\Sigma^{*}$, the configuration vector
 $\DFAConfMap{w} \in \DFAConf$
 has exactly one $\true$;
 the state $\DFAstate$ whose entry is $\true$ is called the \emph{$w$-successor} of $\Autom$.
 $w$ is \emph{accepted} by $\Autom$ if $\DFAWeightMap{w} = \true$. 
\end{definition}

\subsubsection{Recurrent Neural Networks}
Our view of a recurrent neural network (RNN) is almost a black-box. We need only the following two operations: feeding an input word $w\in \Sigma^{*}$ and observing its output (a real number); and additionally, observing the internal state (a vector) after feeding a word. This allows us to model RNNs in the following abstract way. 
\begin{definition}[RNN]\label{def:RNN}
 Let $d\in \N$ be a natural number called a \emph{dimension}.
 A (real-valued) \emph{RNN}  is a triple
\begin{math}
 R=(\alpha_R, \beta_R, g_R)
\end{math},
where $\alpha_R\in \R^{d}$ is an \emph{initial state}, 
$\beta_R\colon \R^d\to \R$ is an \emph{output function}, and $g_R\colon \R^d\times \Sigma \to \R^d$ is called a \emph{transition function}. The set $\R^{d}$ is called a \emph{state space}. 
\end{definition}
\begin{definition}[RNN configuration $\delta_{R}(w)$, output $f_{R}(w)$]\label{def:RNNConfigWeight}
 Let $R$ be the RNN in Def.~\ref{def:RNN}. The transition function $g_R$  naturally extends to words as follows: $g^{*}_R\colon \R^d\times \Sigma^* \to \R^d$, defined inductively by
$ g^{*}_R(x, \varepsilon) = x$ and
$
  g^{*}_R(x, w\sigma) = g_R\bigl(g^{*}_R(x, w), \sigma\bigr)
$, where $w\in \Sigma^{*}$ and $\sigma\in \Sigma$.

The \emph{configuration} $\delta_{R}(w)$ of the RNN $R$ at a word $w$ is defined by 
$\delta_R(w) = g^{*}_{R}(\alpha_R, w)$. The \emph{output} $f_{R}(w)\in \R$, of $R$ for the input $w$, is defined by $f_{R}(w)=\beta_{R}\bigl(\delta_R(w)\bigr)$.
\end{definition}





\subsection{Angluin's \lstar~Algorithm}
\label{subsec:angluin}
Angluin's \lstar-algorithm learns a given DFA $B$ by a series of  \emph{membership} and \emph{equivalence} queries. We sketch the algorithm; see~\cite{DBLP:journals/iandc/Angluin87} for details.
Its outline is in Fig.~\ref{fig:overviewOfLStar}.

  \begin{figure}[tbp]
   \centering
   \centering
 \scalebox{.85}{\begin{tikzpicture}[auto, semithick,remember picture,
    block/.style={rectangle, draw,
        minimum width=5em, text centered, rounded corners, minimum
        height=3em,text width=7em},
    nrblock/.style={rectangle, draw,
        minimum width=5em, text centered, rounded corners, minimum
        height=3em,text width=5em}
    ]
    \node[nrblock](n11){membership query $f_{B}(w)={?}$};
		\node[nrblock,right= 7.3em of n11](n12){extend the table $T$};
		\node[block,below= 3em of n12,text width = 9em](n22){equivalence query $A_{T}\cong^{?} B$};
		\draw[->,thick] (n11) edge[bend left=-10] node [auto=right] {$\bigl(w,f_{B}(w)\bigr)$} (n12);
		\draw[->,thick] (n12) edge[bend left=-10] node [auto=right] {\parbox{6em}{$T$ is not closed\\$\Rightarrow$ pick new $w$}}
		(n11);		
		\draw[->,thick] (n12) edge[bend left=0] node
		{\parbox{8.5em}{$T$ is  \emph{closed} $\Rightarrow$\\  $T$ induces a DFA $A_{T}$}}
		(n22);		
		\draw[->,thick] (n22) edge[bend left=20] node [anchor=30] {\parbox{10.5em}{\hfill no\phantom{hoge}\\ $\Rightarrow$ use a counterexample ($w$ s.t.\ $f_{A_{T}}(w)\neq f_{B}(w)$) as the next $w$}}
		(n11);		
  		\node[right= 2em of n22](n23) {done};
		\coordinate[above= 1.5 em of n11](n01);
		\draw[->,thick] (n01) -- (n11);
		\draw[->,thick] (n22) edge node {yes} (n23);
\end{tikzpicture}
}

\vspace*{-1em}
 \caption{An outline of Angluin's \lstar~algorithm. The target DFA is $B$; a table $T$ gets gradually extended, yielding a DFA $A_{T}$ when it is closed. See also Fig.~\ref{fig:exampleTable}}
   \label{fig:overviewOfLStar}
  \end{figure}
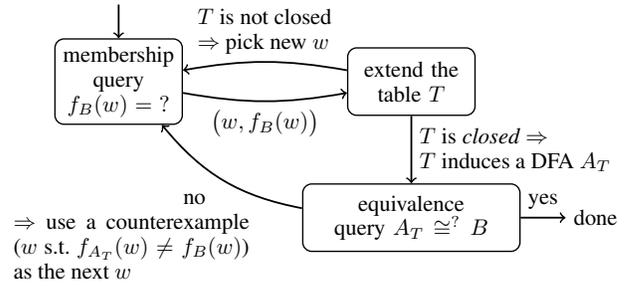

\begin{figure}[t]
\begin{subfigure}{.48\linewidth}
 \centering
 \scalebox{.8}{\begin{math}
 \begin{array}{l||l|l|l}
  \mathcal{A}\backslash \mathcal{T} &\varepsilon & 0  &1 \\\hline
 \varepsilon & \true & \false & \false \\
 0 & \false & \true&\false\\
 \multicolumn{1}{c||}{\vdots} &  
 \multicolumn{1}{c|}{\vdots} &  
 \multicolumn{1}{c|}{\vdots} &  
 \multicolumn{1}{c}{\vdots} \\
 011 & \false & \true & \false
 \end{array}
 \end{math}}
 \caption{A table $T$ for DFA learning
}\label{fig:exampleTable}
\end{subfigure}
 \hfill
\begin{subfigure}{.48\linewidth}
\centering
\scalebox{.8}{\begin{math}
 \begin{array}{l||l|l|l}
  \mathcal{A}\backslash \mathcal{T} &\varepsilon & 0  &1 \\\hline
 \varepsilon & 0.5 & 0 & 0.5 \\
 0 & 0 & 1& 0
\\
 \multicolumn{1}{c||}{\vdots} &  
 \multicolumn{1}{c|}{\vdots} &  
 \multicolumn{1}{c|}{\vdots} &  
 \multicolumn{1}{c}{\vdots} \\
 001 & 0.2 & 0.4 & 0.4
 \end{array}
\end{math}}
\caption{A table $T$ for WFA learning
}\label{fig:exampleTableWeighted}
 \end{subfigure}
\caption{Observation tables for \lstar-style algorithms}
\end{figure}
A \emph{membership query} is a black-box observation of the DFA $B$: it feeds $B$ with a word $w\in\Sigma^{*}$; and obtains $f_{B}(w)\in\{\true,\false\}$, i.e., whether $w$ is accepted by $B$. 

The core of the algorithm is to construct the \emph{observation table} $T$; see Fig.~\ref{fig:exampleTable}.
The table has words as the row and column labels;  its entries are either $\true$ or $\false$. The  row labels are called \emph{access words};  the column labels are \emph{test words}. We let $\mathcal{A},\mathcal{T}$ stand for the sets of access and test words. 
The entry of $T$ at row $u\in\mathcal{A}$ and column $v\in\mathcal{T}$ is given by $f_{B}(uv)$---a value that we can  obtain from a suitable membership query. 

Therefore we  extend a table $T$ by a series of membership queries. We do so until $T$ becomes closed; this is the  top loop in Fig.~\ref{fig:overviewOfLStar}. A table $T$ is \emph{closed} if, for any access word $u\in\mathcal{A}$ and $\sigma\in\Sigma$, there is an access word $u'\in\mathcal{A}$ such that%
\begin{equation}\label{eq:myhillNerodeAngluin}
 f_{B}(u\sigma\, v) =  f_{B}(u' v)
 \quad\text{for each test word $v\in\mathcal{T}$.}
\end{equation}
The closedness condition essentially says that the role of the extended word $u\sigma$ is already covered by some existing word $u'\in\mathcal{A}$. The notion of ``role''  here, formalized in~(\ref{eq:myhillNerodeAngluin}), is a restriction of the well-known Myhill--Nerode relation, from all words $v\in\Sigma^{*}$ to $v\in\mathcal{T}$. 

A closed table $T$ induces a DFA $A_{T}$ (Fig.~\ref{fig:overviewOfLStar}), much like in the Myhill--Nerode theorem. We note that the resulting DFA $A_{T}$ is necessarily minimized. The DFA $A_{T}$ undergoes an \emph{equivalence query} that asks if $A_{T}\cong B$; an equivalence query is answered with a counterexample---i.e., $w\in \Sigma^{*}$ such that $f_{A_{T}}(w)\neq f_{B}(w)$---if $A_{T}\not\cong B$.

 The \lstar~algorithm is a \emph{deterministic} learning algorithm (at least in its original form), unlike many recent learning algorithms that are statistical. 
The greatest challenge in practical use of the \lstar~algorithm  is to answer  equivalence queries. When $B$ is a finite automaton that generates a regular language, there is a complete algorithm for deciding the language equivalence  $A_{T}\cong B$. However, if we use a more expressive model in place of a DFA $B$, checking $A_{T}\cong B$ becomes a nontrivial task.


\subsection{\lstar~Algorithm for WFA Learning}
\label{subsec:balle}
The classic \lstar~algorithm for learning DFAs has seen a \emph{weighted} extension~\cite{DBLP:conf/cai/BalleM15}: it learns a WFA $B$, again via a series of membership and equivalence queries. The overall structure of the WFA learning algorithm stays the same as in Fig.~\ref{fig:overviewOfLStar}; here we highlight major differences. 

Firstly, the entries of an observation table $T$ are now real numbers, reflecting the fact that the value $f_{B}(uv)$ for a WFA $B$ is in $\R$ instead of in $\{\true,\false\}$ (see Def.~\ref{def:WFAWeightOfAWord}). An example of an observation table is given in Fig.~\ref{fig:exampleTableWeighted}.

Secondly, the notion of closedness is adapted to the weighted (i.e., linear-algebraic) setting, as follows. 
A table $T$ is \emph{closed} if, for any access word $u\in\mathcal{A}$ and $\sigma\in\Sigma$, 
 the vector
 \begin{math}
 \bigl(\,f_{B}(u\sigma\, v) \,\bigr)_{v\in \mathcal{T}}
 \in \R^{|T|}
 \end{math}
 can be expressed as a linear combination of the vectors in
 \begin{math}
 \bigl\{\, \bigl(\,f_{B}(u' v) \,\bigr)_{v\in \mathcal{T}}\,\big|\, u'\in \mathcal{A}\bigr\}
 \end{math}.
Note that the vector $\bigl(\,f_{B}(u' v) \,\bigr)_{v\in \mathcal{T}}^{\top}$ in the latter set is precisely the row vector in $T$ at row $u'$.

For example, the table $T$ in Fig.~\ref{fig:exampleTableWeighted} is obviously closed, since the three row vectors are linearly independent and thus span the whole $\R^{3}$. The above definition of closedness comes natural in view of Def.~\ref{def:WFAConfig}. For a WFA, a configuration (during its execution) is not a single state, but a \emph{weighted superposition}  $\WFAconf \in \WFAConf$ of states. The closedness condition asserts that the role of $u\sigma$ is covered by a suitable superposition of words $u'\in\mathcal{A}$. The construction of the WFA $A_{T}$ from a closed table $T$ (see Fig.~\ref{fig:overviewOfLStar}) reflects this intuition. See~\cite{DBLP:conf/cai/BalleM15}. We note that the resulting $A_{T}$ is minimal, much like in~\S{}\ref{subsec:angluin}. 

In the literature~\cite{DBLP:conf/cai/BalleM15}, an observation table $T$ is presented as a so-called \emph{Hankel} matrix. This opens the way to further extensions of the method, such as an approximate learning algorithm via the singular-value decomposition (SVD). 

\section{WFA Extraction from an RNN}\label{sec:ourAlgorithm}
\TODO{The column width exceeds the limit because of the procedure definition!!!}
\begin{figure}[tbp]
  \scalebox{.75}{\begin{tikzpicture}[auto, semithick,remember picture,
    block/.style={rectangle, draw,
        minimum width=5em, text centered, rounded corners, minimum
        height=3em,text width=7em},
    rnnblock/.style={rectangle, draw,
        minimum width=5em, text centered, rounded corners, minimum
        height=3em,text width=11.3em},
    remblock/.style={rectangle, draw, dashed,
        minimum width=5em, text centered, rounded corners, minimum
        height=3em,text width=21em},
    wfablock/.style={rectangle, draw,
        minimum width=5em, text centered, rounded corners, minimum
        height=3em,text width=14em}
    ]
		 \node[rnnblock](n11){
		 \begin{minipage}{10.3em}
		  \underline{RNN $R$}
		  

			 \noindent
			  ${\color{red}\R^{d}\color{black}}$:
			  \parbox[t]{8.3em}{state space = \\ configuration space}
		 \end{minipage}
		 };
		 \node[wfablock,right= 3em of n11](n12){
		 \begin{minipage}{13em}
			 \underline{WFA $M$}

		         \begin{math}
			  \begin{array}{rl}
			   Q_{A}:&\text{state space} \\
			   {\color{blue}\R^{Q_{A}}\color{black}}:&\text{configuration space}
			  \end{array}
			 \end{math}

		 \end{minipage}
		 };
		 \draw[->,thick] (n11) edge[bend left=-0] node
		 [auto=right] (extr) {extract} (n12);
		 \node[remblock,below=1.7em of extr] (rem1) {configuration abstraction by
		 $p\colon {\color{red}\R^{d}\color{black}}\to {\color{blue}\R^{Q_{A}}\color{black}}$};
		 \draw[->,dashed,thick] (rem1) -- (extr);
\end{tikzpicture}
}\caption{An outline of our WFA extraction }
\label{fig:WFAExtractionAndConfigSpaceAbstraction}
\end{figure}
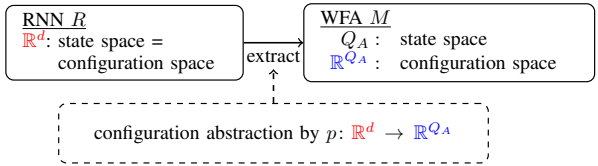
We present our main contribution, namely a procedure that extracts a
WFA from a given RNN. 
After briefly describing its outline (that is the same as Fig.~\ref{fig:overviewOfLStar}), we focus on the greatest challenge of answering equivalence queries.

\subsection{Procedure Outline}
Our procedure uses the weighted  \lstar~algorithm sketched in~\S{}\ref{subsec:balle}. As we discussed in~\S{}\ref{subsec:angluin}, the greatest challenge  is how to answer equivalence queries; our novel approach is to use \emph{regression} and synthesize what we call a \emph{configuration abstraction function} $p\colon \R^{d}\to \R^{Q_{A}}$. See Fig.~\ref{fig:WFAExtractionAndConfigSpaceAbstraction}.


The outline of our procedure thus stays the same as in
Fig.~\ref{fig:overviewOfLStar}, but we need to take care about noisy
outputs from an RNN because they prevent the observation table from being
precisely closed (in the sense of \S{}\ref{subsec:balle}).  To resolve this issue,
we use a noise-tolerant algorithm~\cite{DBLP:conf/cai/BalleM15} which approximately determines
whether the observation table is closed.
This approximate algorithm employs SVD and cuts off singular values that are smaller than a threshold called \emph{rank tolerance}.
In the choice of a rank tolerance, we face the trade off between accuracy and regularity.
If the rank tolerance is large, the WFA learning algorithm tends to ignore the counterexample given by an equivalence query and results in producing an inaccurate WFA.
We use a heuristic to decrease the rank tolerance when two or more equivalence queries return the same counterexample. See \ArxivVersion{Appendix~\ref{appendix:rankTolerance}}\CameraReadyVersion{Appendix~A.1 of~\cite{DBLP:journals/corr/abs-1904-02931}} for details. 

\subsection{Equivalence Queries for WFAs and RNNs}
\label{subsec:equery}
\begin{algorithm}[tbp]
    \caption{Answering equivalence queries}
    \label{alg:equivalence}
    \label{alg:aaai_bfs}
  \begin{algorithmic}[1]
  
  \Procedure{Ans-EqQ}{}
  \Statex\label{line:inputEqQuery} {\bfseries Input:}  RNN $R=(\alpha_R, \beta_R, g_R)$,
  WFA $A=(Q_A, \alpha_A, \beta_A, (A_\sigma)_{\sigma\in \Sigma})$, 
  error tolerance  $e >0$ and  concentration threshold $M\in\mathbb{N}$
  \Statex {\bfseries Output:} a counterexample, or $\mathtt{Equivalent}$
  \State{Initialize $p\colon \R^{d}\to \R^{Q_A}$ so that $p(\delta_R(\varepsilon)) = \delta_A(\varepsilon)$}
  \State{$\mathtt{queue} \gets \mylist{\varepsilon}$;\;$\mathtt{visited} \leftarrow \emptyset$} 
  \label{line:queueInitialization}
  \While{$\mathtt{queue}$ is non-empty} \label{line:whileLoopQueue}
   \State{$h\leftarrow \mathbf{pop}(\mathtt{queue})$}\par
   \Comment{Pop the element of the maximum priority}
  \label{line:popFromQueue}
  \If{$\abs{f_R(h) - f_A(h)} \ge e$}
  \label{line:prototype1}
  \Return{$h$}
  \Comment{return a counterexample}
  \label{line:prototype2}
  \EndIf
  \State{$\mathtt{result}\leftarrow \textsc{Consistent?}(h, \mathtt{visited}, p)$}
  \label{line:refinePStart}
  \label{line:consistencyCheckIsCalled}
  \If{$\mathtt{result} = \mathtt{NG}$}
  \State\label{line:relearn} 
\parbox[t]{.35\textwidth}{  learn $p$ by regression, so that
  $p(\delta_R(h')) = \delta_A(h')$ holds for all $h'\in \mathtt{visited}\cup\{h\}$
}
  \EndIf
  \label{line:refinePEnd}
  \State{$\mathtt{visited} \leftarrow \mathtt{visited}\cup \set{h}$}
  \label{line:addToVisited}
  \State{$\mathtt{visited'} \leftarrow p(\delta_R(\mathtt{visited}))$}
   \State{$\mathit{\#vn} \gets \left|
        \set{x\in \mathtt{visited'} \mid x\simeq_A p(\delta_R(h))}
      \right|$}\label{line:useOfEta}
   \If{$\mathit{\#vn} \le M$}
   \State{$\displaystyle \mathsf{pr} \gets \min_{h'\in \mathtt{visited}\setminus{}\{h\}} d(p(\delta_R(h)), p(\delta_R(h')))$}\par
   \Comment{$d$ is the Euclidean distance}
   \label{line:calcPriority}
   \State{\textbf{push} $h\sigma$ \textbf{to} $\mathtt{queue}$ with priority $\mathsf{pr}$ \textbf{for}\;$\sigma\in \Sigma$}\label{line:pushToQueue}
   \EndIf
  \EndWhile
  \Return{ $\mathtt{Equivalent}$ }
  \label{line:returnEquivalentAfterLoop}
  \EndProcedure
  \end{algorithmic}
  \end{algorithm}

Algorithm~\ref{alg:equivalence} shows our procedure to answer an equivalence query.
The procedure $\textsc{Ans-EqQ}$ is the main procedure, and it returns either $\mathtt{Equivalent}$ or a counterexample word (as in Fig.~\ref{fig:overviewOfLStar}). 
It calls the auxiliary procedure $\textsc{Consistent?}$, which decides if we refine the current configuration abstraction function $p\colon \R^{d}\to\R^{Q_{A}}$ in Fig.~\ref{fig:WFAExtractionAndConfigSpaceAbstraction} (Line~\ref{line:relearn}). 

\subsubsection{Best-First Search for a Counterexample}

The procedure $\textsc{Ans-EqQ}$ is essentially a best-first search for a \emph{counterexample}, that is, a word $h\in\Sigma^{*}$ such that the difference of the output values $f_{R}(h)$ from the RNN and $f_{A}(h)$ from the WFA is larger than \emph{error tolerance} $e (>0)$.
We first outline $\textsc{Ans-EqQ}$ and then go into the technical detail.

We manage counterexample candidates by the priority queue $\mathtt{queue}$, which
gives a higher priority to a candidate more likely to be a counterexample.
The already investigated words are in the set $\mathtt{visited}$.
The queue $\mathtt{queue}$
initially contains only the empty word $\varepsilon$
(Line~\ref{line:queueInitialization}).

We search for a counterexample in the main loop starting from
Line~\ref{line:whileLoopQueue}.  Let $h$ be a word popped from $\mathtt{queue}$,
that is, the candidate most likely to be a proper counterexample among the words in the queue.
If $h$ is a counterexample, $\textsc{Ans-EqQ}$ returns it
(Lines~\ref{line:prototype1}--\ref{line:prototype2}).  Otherwise, after refining
the configuration abstraction function $p$
(Lines~\ref{line:refinePStart}--\ref{line:refinePEnd}), new candidates
$h\sigma$, the extension of $h$ with character $\sigma$, are pushed to
$\mathtt{queue}$ with their priorities \emph{only if} the neighborhood of $h$ in
the state space of the WFA $A$ does not contain sufficiently many already
investigated words---i.e., only if it has not been investigated sufficiently
(Lines~\ref{line:useOfEta}--\ref{line:pushToQueue}).  This is because, if the
neighborhood of $h$ has been investigated sufficiently, we expect that
the neighborhoods of the new candidates $h\sigma$ have also been investigated and,
therefore, that the words $h\sigma$ do not have to be investigated furthermore. 
We use $p$: 1) to decide if the neighborhood of $h$ has been investigated sufficiently (Line~\ref{line:useOfEta}); and 
2) to calculate the priorities of the new candidates (Line~\ref{line:calcPriority}).
Note that we add $h$ to $\mathtt{visited}$ in Line~\ref{line:addToVisited} since it has been
investigated there.
If all the candidates are not a counterexample, \textsc{Ans-EqQ}
returns $\mathtt{Equivalent}$ (Line~\ref{line:returnEquivalentAfterLoop}).

\subsubsection{Configuration Abstraction Function $p$}


To use $p$ for the above purposes, the property we expect from the configuration abstraction function $p\colon \R^{d}\to \R^{Q_{A}}$ is as follows:
\begin{equation}\label{eq:desiredPropertyForconfigurationabstractionfunction}
 p(\delta_{R}(h)) \approx \delta_{A}(h)
 \quad\text{for as many $h\in\Sigma^{*}$ as possible.}
\end{equation}
See Def.~\ref{def:WFAConfig} and~\ref{def:RNNConfigWeight}
for $\delta_{A}\colon \Sigma^{*}\to \R^{Q_{A}}$ and $\delta_{R}\colon \Sigma^{*}\to \R^{d}$, respectively.
To synthesize such a function $p$, we employ \emph{regression} using the data $\bigl\{\,\bigl(\delta_{R}(h'), \delta_{A}(h')\bigr)\in \R^{d}\times\R^{Q_{A}}\,\big|\,h\in\mathtt{visited}\,\bigr\}$.  See Line~\ref{line:relearn}. 
Note that we can use any regression method to learn $p$.

We refine $p$ during the best-first counterexample search. Specifically, in Line~\ref{line:consistencyCheckIsCalled}, we use the procedure $\textsc{Consistent?}$ to check if the current $p$---obtained by regression---is consistent with a counterexample candidate $h$. The consistency means that $p(\delta_{R}(h))$ and $\delta_{A}(h)$ are close to each other, which is formalized by the relation $\simeq_A$ defined later.  If the check fails (i.e., if  $\textsc{Consistent?}$ returns $\mathtt{NG}$), we refine $p$ by regression to make $p$ consistent with $h$ (and the already investigated words in $\mathtt{visited}$). 
See Line~\ref{line:relearn}.

\subsubsection{Consistency Checking by $\textsc{Consistent?}$}
The procedure $\textsc{Consistent?}$ in Line~\ref{line:consistencyCheckIsCalled} is defined as follows: it returns $\mathtt{NG}$ if there exists $ h'\in \mathtt{visited}$ such that
\begin{equation}
 \delta_A(h') \not\simeq_A p(\delta_R(h')) \text{ and }p(\delta_R(h'))\simeq_A p(\delta_R(h)),
  \label{eq:consistencyCheck}
\end{equation}
and returns $\mathtt{OK}$ otherwise.
The basic idea of $\textsc{Consistent?}$ is to return $\mathtt{NG}$ if
$p(\delta_R(h))\not\simeq_A\delta_A(h)$ because it means the violation of the
desired property (\ref{eq:desiredPropertyForconfigurationabstractionfunction}).
However, to reduce the run-time cost of refining $p$ and to prevent learning
from outliers, we adopt the alternative approach presented above, which is taken from
\citet{DBLP:conf/icml/WeissGY18}.

The existence of $h'$ satisfying the condition (\ref{eq:consistencyCheck}) approximates
the violation of the property (\ref{eq:desiredPropertyForconfigurationabstractionfunction})
in the following sense. If there is a word $h'$ satisfying the first part of the condition (\ref{eq:consistencyCheck}), $p$ has to be refined because we find
the property (\ref{eq:desiredPropertyForconfigurationabstractionfunction}) violated with $h'$.
The second part of the condition (\ref{eq:consistencyCheck}) means that $h'$ seems to behave similarly to $h$ according to the configuration abstraction function $p$.  We expect the second part to prevent $p$ from being refined with outliers because the neighborhoods of the words used for refining $p$ must have been investigated twice or more.

\subsubsection{Equivalence Relation $\simeq_A$}
For a given WFA $A$,
we define the relation $\simeq_A$ in the configuration space $\WFAConf{}$ by
\begin{align}
	\vec x \simeq_A \vec y
	\iff
	\textstyle\sum_{i=1}^{\abs{Q_A}} \beta_i^2 (x_i-y_i)^2 < \frac{e^2}{\abs{Q_A}},
\end{align}
where $\beta$ is the final vector of the WFA $A$.
It satisfies the following.
\begin{enumerate}
	\item If $x\simeq_A y$ holds, the difference of the output values 
	for the configurations $x$ and $y$ of the WFA $A$ is smaller than the error tolerance $e$, that is,
	$\abs{(x-y)\cdot \beta} < e$.
	\item If the $i$-th element of the final vector $\beta$ becomes large, the neighborhood $\set{y\in \R^{Q_A} \mid x\simeq_A y}$ of $x$ shrinks in the direction of the $i$-th axis.
	\item The neighborhood defined by $\simeq_A$ is an ellipsoid---a reasonable  variant of an open ball as a neighborhood.
\end{enumerate}

\subsubsection{A Heuristic for Equivalence Checking of a WFA and an RNN}
Although the best-first search above works well, we introduce an additional heuristic to
improve the run-time performance of our algorithm furthermore.  The heuristic
deems $R$ and $A$ to be equivalent if word $h$ previously popped from
$\mathtt{queue}$ is so long that it is impossible to occur in the training of the RNN.  This heuristic is based on
the expectation that, when an impossible word is the most likely to be a
counterexample, all possible words are unlikely to be a counterexample, and so
$R$ and $A$ are considered to be equivalent.  This heuristic is adopted immediately
after popping $h$ (Line~\ref{line:popFromQueue}), as follows: we suppose
that the maximum length $L$ of possible words is given; and, if the length of
$h$ is larger than $L$, \textsc{Ans-EqQ} returns $\mathtt{Equivalent}$.  We
confirm the usefulness of this heuristic in \S{}\ref{sec:implExpr} empirically.

\subsubsection{Termination of the Procedure}
Algorithm~\ref{alg:equivalence} does not always terminate in a finite amount of time.  If the procedure does not find any counterexample at Line~\ref{line:prototype2} and the points $p(\delta_R(\mathtt{visited}))$ are so scattered in the configuration space that the value $\#vn$ at Line~\ref{line:useOfEta} is always small, words are always pushed to $\mathtt{queue}$ at Line~\ref{line:pushToQueue}. In that case, the condition to exit the main loop at Line~\ref{line:whileLoopQueue} is never satisfied.

\subsection{Comparison with Weiss et al., 2018}\label{subsec:comparisonWithWeissEtAl}
Our WFA extraction method can be seen as a weighted extension of the \lstar-based procedure~\cite{DBLP:conf/icml/WeissGY18} to extract a DFA from an RNN.  Note that a WFA defines a function of type $\Sigma^*\to \R$ and a DFA defines a function of type $\Sigma^* \to \set{\true, \false}$.

The main technical novelty of the method in~\cite{DBLP:conf/icml/WeissGY18} is how to answer equivalence queries.
It features the \emph{clustering} of the state space $\R^{d}$ of an RNN into finitely many clusters, using support-vector machines (SVMs).
Each cluster of $\R^{d}$ is associated with a state $q\in Q_{A}$ of the DFA.

Our theoretical observation is that such clustering amounts to giving a function $p\colon \R^{d}\to Q_{A}$.
Moreover, for a DFA, $Q_{A}$ is the configuration space (as well as the state space, see Def.~\ref{def:DFA}).
Therefore, our WFA extraction method can be seen as an extension of the DFA extraction procedure in~\cite{DBLP:conf/icml/WeissGY18}.


\section{Experiments}\label{sec:implExpr}
\taro{Algorithm or procedure? $\to$ procedure! (Sep 5 22:32)}

We conducted experiments to evaluate the utility of our 
regression-based WFA extraction method. Specifically, we pose the following questions.
\begin{description}
 \item[RQ1] Do the WFAs extracted by our algorithm approximate the original RNNs
            accurately?
            %
 \item[RQ2] Does our algorithm work well with RNNs whose 
            expressivity is beyond WFAs?
            %
 \item[RQ3] Do the extracted WFAs run more efficiently, specifically in inference time, than given RNNs?
\end{description}
For \textbf{RQ1}, we compared the performance with a baseline algorithm (a straightforward adaptation of \citet{DBLP:conf/cai/BalleM15}'s \lstar algorithm). Here we focused on ``automata-like'' RNNs, that is, those RNNs trained from an original WFA $A^{\bullet}$. For \textbf{RQ2}, we used an RNN that exhibits ``context-free'' behaviors.


\noindent\textbf{Experimental Setting}
We implemented our method in Python.
We write \regr{n} for the
algorithm where the concentration threshold $M$ in $\textsc{Ans-EqQ}$ is set to
$n$; other parameters are fixed as follows: error tolerance $e = 0.05$ and
heuristic parameter $L=20$.  We adopt the
Gaussian process regression (GPR) provided by scikit-learn as a
configuration abstraction function $p$ (we also tried the kernel ridge
regression but  GPR worked empirically better).  Throughout the experiments,
our RNNs are 2-layer LSTM networks with dimension size 50, implemented by
TensorFlow.  The experiments were on a
g3s.xlarge instance on Amazon Web Service (May 2019), with a NVIDIA Tesla M60
GPU, 4 vCPUs of Xeon E5-2686 v4 (Broadwell), and 8GiB memory.

\begin{table*}[t]
\newcommand{\tbcolor}{\cellcolor{green!25}}
 \caption{Experiment results, where we extracted a WFA $A$ from an RNN $R$ that is trained to mimic the original WFA $A^{\bullet}$.
In each cell ``n/m'',
 ``n'' denotes the average of  MSEs between $A$ and $R$ (the unit is $10^{-4}$), taken over five random WFAs $A^{\bullet}$ of the designated alphabet size $|\Sigma|$ and the state-space size $|Q_{A^{\bullet}}|$. ``m'' denotes the average running time (the unit is second).  The ``Total'' row describes the average over all the experiment settings. The highlighted cell designates the best performer in terms of errors. Timeout was set at 10,000 sec. $\regr{2\text{--}5}$ are our regression-based methods; $\search{500\text{--}5000}$ are the baseline.  $\search{5000}$ is added to compare the accuracy when the running time is much longer.}
\parbox[c]{.2\textwidth}{
Results for  ``more WFA-like'' RNNs $R$, i.e., those RNNs $R$ which are trained from WFAs $A^{\bullet}$ using a uniformly sampled data set $T\subseteq \Sigma^{*}$ 
}\qquad
\scalebox{.75}{
  \begin{tabular}{c||c|c||c|c|c|c|c}
  $(|\Sigma|, |Q_{A^{\bullet}}|)$        & \regr{2}     & \regr{5}     & \search{500} & \search{1000} & \search{2000} & \search{3000} &  \search{5000}  \\ \hline \hline
  $(4,10)$  & \tbcolor 2.17 / 286   &  2.39 / 338   & 26.8 / 165   & 9.77 / 279    & 4.36 / 545    & 4.07 / 716 & 2.33 / 1390 \\
  $(6,10)$  & 2.45 / 1787  & 2.54 / 1302  & 6.99 / 386   & 4.48 / 641    & 4.08 / 1218   & 3.15 / 1410 & \tbcolor 2.28 / 2480 \\
  $(10,10)$ & 4.68 / 7462  &  4.46 / 5311  & 22.5 / 928   & 11.9 / 1562   & 5.90 / 3521   & 4.55 / 3638 & \tbcolor 3.55 / 5571 \\
  $(10,15)$ &  5.62 / 8941  & 5.78 / 8564  & 21.2 / 2155  & 10.6 / 4750   & 7.87 / 5692   & 5.71 / 7344 & \tbcolor 5.27 / 7612 \\
  $(10,20)$ &  3.70 / 7610  & 3.79 / 7799  & 6.24 / 2465  & 10.1 / 2188   & 6.13 / 3106   & 3.70 / 5729 & \tbcolor 3.63 / 7473 \\
  $(15,10)$ & 7.34 / 9569  & \tbcolor 5.52 / 10000 & 13.5 / 3227  & 8.01 / 6765   & 6.07 / 7916   & 5.98 / 8911 & 6.17 / 8979 \\
  $(15,15)$ & 8.44 / 10000 & \tbcolor 5.58 / 9981  & 16.3 / 2675  & 9.24 / 4850   & 7.28 / 5135   & 9.88 / 7204 & 6.44 / 8425 \\
  $(15,20)$ & 9.16 / 7344  & 5.15 / 7857  & 13.7 / 2224  & 7.26 / 3823   & 6.60 / 5744   & 4.96 / 5674 & \tbcolor 4.01 / 9464 \\ \hline 
  Total     & 5.45 / 6625  &  4.40 / 6394  & 15.9 / 1778  & 8.92 / 3107   & 6.04 / 4110   & 5.25 / 5078 & \tbcolor 4.21 / 6549 \\
 \end{tabular}
}
\\[.2em]
\parbox[c]{.2\textwidth}{
Results for ``realistic'' RNNs $R$ with \emph{nonuniform input domains}, i.e., those $R$ which  are trained from WFAs $A^{\bullet}$ using a data set $T\subseteq\Sigma^{*}$ in which some specific patterns are prohibited

}\qquad
\scalebox{.75}{
 \begin{tabular}{c||c|c||c|c|c|c|c}
  $(|\Sigma|, |Q_{A^{\bullet}}|)$ & \regr{2}    & \regr{5}    & \search{500} & \search{1000} & \search{2000} & \search{3000} & \search{5000} \\ \hline \hline
  $(4,10)$            & 7.73 / 696  & 7.07 / 1135 & 15.0 / 199   & 7.96 / 424    & 6.62 / 650    & \tbcolor 6.61 / 762 & 9.06 / 1693 \\
  $(6,10)$            & 4.92 / 1442 & 7.43 / 1247 & \tbcolor 1.46 / 552   & 6.95 / 660    & 5.90 / 1217   & 8.78 / 1557 & 3.54 / 2237 \\
  $(10,10)$           & 5.02 / 5536 & 4.28 / 5951 & 7.70 / 1117  & 11.0 / 1738   & 4.77 / 2635   & \tbcolor 3.52 / 3926 & 4.52 / 4777 \\
  $(10,15)$           & 7.15 / 6977 & \tbcolor 4.35 / 8315 & 19.4 / 1552  & 13.8 / 3271   & 16.8 / 3209   & 8.57 / 5293 & 5.08 / 6522 \\
  $(10,20)$           & \tbcolor 6.98 / 4697 & 8.06 / 6704 & 18.6 / 1465  & 11.8 / 2046   & 12.7 / 2851   & 9.03 / 4259 & 8.01 / 4856 \\
  $(15,10)$           & \tbcolor  5.97 / 8747 & 6.77 / 8882 & 23.3 / 2359  & 11.2 / 4668   & 9.88 / 6186   & 6.24 / 7557 & 6.02 / 8245 \\
  $(15,15)$           & \tbcolor  5.78 / 8325 & 8.71 / 7546 & 16.6 / 2874  & 7.31 / 4380   & 9.92 / 6015   & 9.89 / 7110 & 6.40 / 8358 \\
  $(15,20)$           & \tbcolor  4.60 / 7652 & 8.56 / 8334 & 36.9 / 1893  & 23.7 / 3069   & 12.8 / 3987   & 12.0 / 5262 & 8.38 / 6441 \\ \hline 
  Total               & \tbcolor 6.02 / 5510 & 6.90 / 6015 & 19.0 / 1502  & 11.7 / 2532   & 9.92 / 3344   & 8.08 / 4466 & 6.38 / 5391 \\
\end{tabular}
}
 \label{tbl:rq1-result}
\end{table*}

\begin{figure*}[tbp]
\begin{minipage}{0.25\linewidth}
 \centering
  \includegraphics[width=\textwidth]{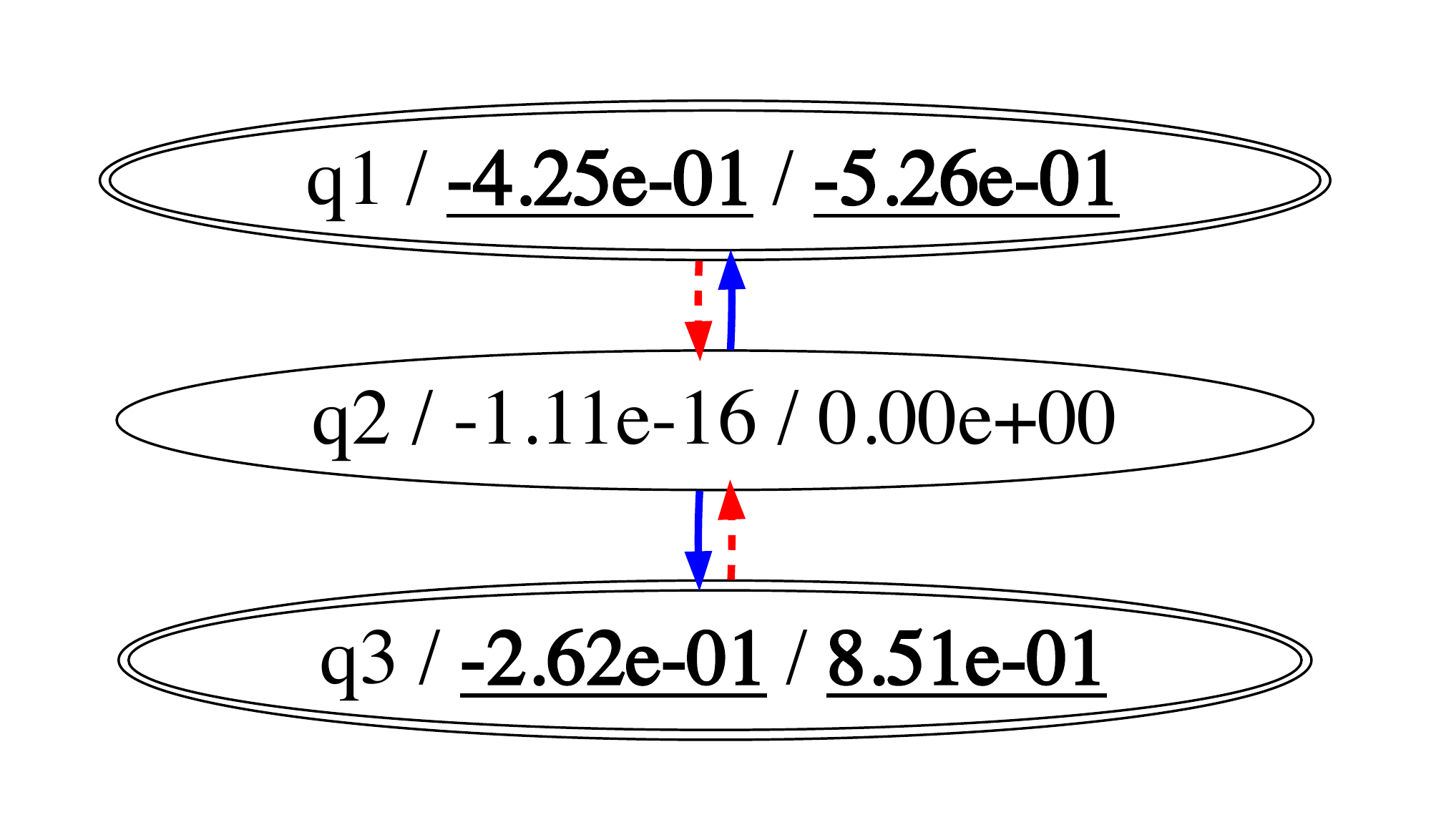} 
\end{minipage}
 \hfill
\begin{minipage}{0.70\linewidth}
 \centering
  \includegraphics[width=\textwidth]{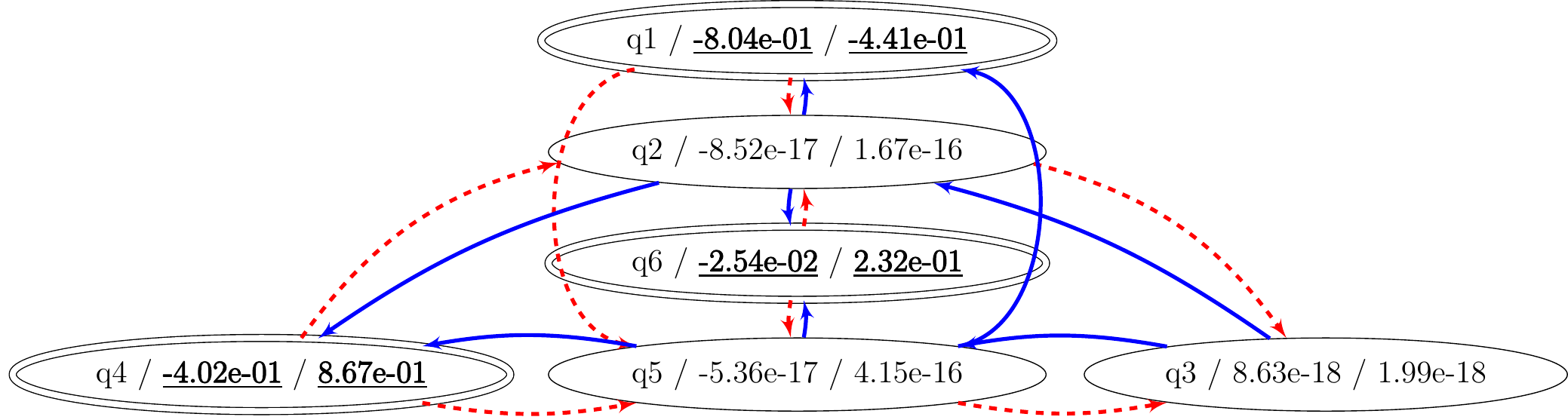} 
\end{minipage}
 \caption{The WFAs extracted by $\regr{5}$ (left) and $\regr{15}$ (right). 
In a state label ``$q/m/n$'', $q$ is the state name, $m$ is the initial value and $n$ is the final value. Bigger values are underlined; other values are negligibly small.
 The dotted and solid edges are labelled with ``\texttt{(}'' and ``\texttt{)}'', respectively; the edges with labels $0,1,\dotsc, 9$ are omitted. The edge weights are omitted for simplicity, too; the weight threshold for showing transitions is $0.01$. \ArxivVersion{Full details on these WFAs are in Appendix~\ref{appendix:wparenWFADetail}.}\CameraReadyVersion{Appendix~B.2 of~\cite{DBLP:journals/corr/abs-1904-02931}.}}
 \label{fig:extracted_WFAs}
\end{figure*}
\subsection{RQ1: Extraction from RNNs Modeling WFAs}\label{subsec:experiment_RQ1}
%
This experiment examines how well our algorithm work for RNNs modeling WFAs.  To
do so, we first train RNNs using randomly generated WFAs; we call those WFAs the
\emph{origins}.  Then, we evaluate our algorithm compared with a baseline from
two points: accuracy of the extracted WFAs against the trained RNNs and running
times of the algorithms.  We report the results after presenting the details of
the baseline, how to train RNNs, and how to evaluate the two algorithms.

\noindent\textbf{The Baseline Algorithm: \search{n}}  As our extraction algorithm,
the baseline algorithm \search{n}, which is parameterized over an integer $n$,
is a straightforward adaptation of \citet{DBLP:conf/cai/BalleM15}'s \lstar algorithm.  The
difference is that equivalence queries in the baseline are implemented in
breath-first search, as follows.  Let $R$
be a given RNN and $A$ be a WFA being constructed.  For each equivalence
query, the baseline searches for a word $w$ such that $|f_A(w)
- f_R(w)| > e$ (where $e= 0.05$), in the breadth-first manner.  If
such a word $w$ is found it is returned as a
counterexample. The search is restricted to the first  $i+n$ words, where $i$ is the index of the counterexample word found in the previous equivalence query. If no counterexample $w$ is found within this search space, 
  the baseline algorithm deems $A$ to be equivalent to
$R$.  Obviously, if $n$ is larger,  more counterexample
candidates are investigated.

\taro{\search{n} is different from the actual implementation.}

\noindent\textbf{Target  RNNs $R$ trained from WFAs $A^{\bullet}$}\quad Table~\ref{tbl:rq1-result} reports the accuracy  of the extracted WFAs $A$, where the target RNNs $R$ are obtained from original WFAs $A^{\bullet}$ in the following manner. Given an alphabet $\Sigma$ of a designated size (the leftmost column), we first
generated a WFA
$A^{\bullet}$ such that 1) the state-space size  $|Q_{A^{\bullet}}|$ is as  designated  (the leftmost column),  and 2)  the initial vector $\alpha_A^{\bullet}$, the final
vector $\beta_A^{\bullet}$, and the transition matrix $A_\sigma^{\bullet}$ are randomly chosen (with normalization so that its outputs are in $[0,1]$). Then we constructed a dataset $T$ by sampling 9000 words
$w$ such that $w \in \Sigma^*$ and $|w| \le 20$; this $T$ was used to train an RNN $R$, on the set
of input-output pairs of $w \in T$ and $f_{A^{\bullet}}(w)$ for 10 epochs.  

A
simple way to sample words $w\in T$ is by the uniform distribution. With $T$ covering the input space of the WFA $A^{\bullet}$ uniformly, we expect the resulting RNN $R$ to inherit properties of $A^{\bullet}$ well.   The top table in Table~\ref{tbl:rq1-result} reports the results in this ``more WFA-like'' setting.  

However, in many applications, the
 input domain of the data used for
training RNNs are nonuniform, sometimes even excluding some
specific patterns. To evaluate our method in such realistic settings,
we  conducted another set of experiments whose results are in the bottom of Table~\ref{tbl:rq1-result}.
Specifically, for training $R$ from $A^{\bullet}$, we used a dataset $T$ that only contains those words  $\sigma_{1}\sigma_{2} \dotsc \sigma_{n}$ which satisfy the following condition: if $\sigma_{i} = \sigma_{j}$ ($i < j$), then
$\sigma_{i} = \sigma_{i+1} = \dotsc = \sigma_{j-1} = \sigma_{j}$. For example, for 
 $\Sigma = \{ a, b, c \}$, $aabccc$ and $baaccc$ may be
in $T$, but $aaba$ may not. 

\noindent\textbf{Evaluation}
In order to evaluate accuracy, we calculated
the mean square error (MSE) of the extracted WFA $A$ against the RNN $R$, using a dataset
$V$ of words sampled from an appropriate distribution, namely the one used in
training the RNN $R$ from $A^{\bullet}$.   The dataset $V$ is sampled so that it does not
overlap with the training dataset $T$ for $R$.  

\noindent\textbf{Results and Discussions}
In the  experiments  in Table~\ref{tbl:rq1-result}, we
considered 8 configurations for generating the original WFA $A^{\bullet}$ (the leftmost column).
The unit of MSEs are $-4$---given also that the outputs of the original WFAs $A^{\bullet}$ are normalized to $[0,1]$, we can say that the MSEs are small enough.


In
the top table in Table~\ref{tbl:rq1-result} (the ``more WFA-like'' setting),
\search{5000} and \regr{5} achieved the first- and second-best performance in
terms of accuracy, respectively (see the ``Total'' row).  More generally, we can
find the trend that, as an extraction runs longer, it performs better.  We
conjecture its reason as follows.  Recall that all the RNNs are trained on words
sampled from the uniform distribution.  This means that all words would be
somewhat informative to approximate the RNNs.  As a result, the performance is
more influenced by the amount of counterexamples---i.e., how long time
extraction takes---than on their ``qualities.''

The exception of this trend is \regr{2}, which took a longer time but performed
worse than \search{3000}, \search{5000}, and \regr{5}. In particular, \regr{2}
performed well for smaller alphabets ($|\Sigma| \in \{ 4, 6, 10 \}$) but not so
when $|\Sigma| = 15$.  The role of the parameter $M$ in $\regr{M}$ (i.e., in
Algorithm~\ref{alg:aaai_bfs}) is a threshold to control how many words
configuration regions of a WFA are investigated with.  Thus, we conjecture that
the use of too small $M$ limits the input space to be investigated excessively,
which is more critical as the input space is larger, eventually biasing the
counterexamples $h$ (in Algorithm~\ref{alg:aaai_bfs}), though the RNNs are
trained on the uniform distribution, and making refinement of WFAs less
effective.



In the bottom table in Table~\ref{tbl:rq1-result} (the ``realistic'' setting), 
\regr{2} performs significantly better than the other (and the best among all the procedures) in terms of accuracy. This is the case even for a large alphabet
($|\Sigma| = 15$).  This indicates that, in the cases that an RNN is trained with a
nonuniform dataset,
making the investigated input space larger by big $M$ could even degrade the accuracy
performance. A possible reason for this degradation is as follows.
 Some words (such as $aba$) are prohibited in the sample set $T$, and
the behaviors of the RNN $R$ for those prohibited words are
unexpected.  Therefore, those prohibited words should not be
useful to refine a WFA.  The use of small $M$ could prevent such meaningless (or even harmful) counterexamples $h$ from being investigated.
This discussion raises another question: how can we find an optimal $M$?  We leave it as future work.

Let us briefly discuss the sizes of the extracted WFAs. The general trend is that the extracted WFAs $A$ have a few times greater number of states than the original WFAs $A^{\bullet}$ used in training $R$. For example, in the setting of the top table in Table~\ref{tbl:rq1-result}, for $|\Sigma|=15$ and $|Q_{A^{\bullet}}|=20$, the average number of the states of the extracted $A$ was 38.2.



\newcommand{\wparen}{\mathtt{wparen}}
\subsection{RQ2:   Expressivity  beyond WFAs}
We conducted experiments to examine
how well our method works for RNNs modeling  languages that cannot be expressed by any WFA.
Specifically, we used an RNN that models the following function $\wparen\colon \Sigma^{*} \to [0,1]$: $\Sigma = \{(,),0,1,...,9\}$,
\begin{math}
 \wparen(w)=  1 - (1/2)^{N}
\end{math}
if all the parentheses in $w$ are balanced
(here $N$ is the depth of the deepest balanced parentheses in  $w$);  and
\begin{math}
 \wparen(w)= 0
\end{math} otherwise.
This $\wparen$
 is a weighted variant of a (non-regular) language of balanced parentheses.
For instance,
 $\wparen(\text{``((3)(7))))''}) = 0$,
 $\wparen(\text{``((3)(7))''}) = 1 - (1/2)^2 = 3/4$, and
 $\wparen(\text{``(a)(b)(c)''}) = 
1/2$.

 We trained an RNN $R$ as follows.
 We generated datasets $T_{\mathrm{good}}$ and $T_{\mathrm{bad}}$,
 and trained an RNN $R$ on the set of input-output pairs of $w \in T_{\mathrm{good}} \cup T_{\mathrm{bad}}$ and $\wparen(w)$.
 The  dataset $T_{\mathrm{good}}$ consists of randomly generated words where all the parentheses are balanced;  $T_{\mathrm{bad}}$ is constructed similarly, except that we apply suitable mutation to each word, which most likely makes the parentheses unbalanced. \ArxivVersion{See Appendix~\ref{appendix:wparenDataGen} for details.}\CameraReadyVersion{See Appendix~B.1 of~\cite{DBLP:journals/corr/abs-1904-02931} for details.}

Fig.~\ref{fig:extracted_WFAs} shows the WFAs extracted from $R$.  Remarkable observations here are as follows.
\begin{itemize}
 \item The shapes of the WFAs---obtained by ignoring clearly negligible weights---give rise to  NFAs that recognize balanced parentheses up-to a certain depth.
 \item As the parameter $M$ in $\regr{M}$ grows, the recognizable depth bound grows: depth one with $\regr{5}$; and depth two with $\regr{15}$. 
\end{itemize}
We believe these observations demonstrate important features, as well as  limitations, of our method. Overall, the extracted WFAs expose interpretable structures hidden in an RNN: the NFA structures in Fig.~\ref{fig:extracted_WFAs} are human-interpretable (they are easily seen to encode bounded balancedness) and machine-processable (such as determinization and minimization). It is also suggested that the parameter $M$ gives us flexibility in the trade-off of extraction cost and accuracy. At the same time, we can only obtain a truncated and regularized version of the RNN structure---this is an inevitable limitation as long as we use the  formalism of WFAs.



We also note that, in each of the two extracted WFAs,  the transition matrices $A_{\sigma}$ are similar for all  $\sigma \in \{0,1,\dotsc,9\}$ (the entries at the same position have the same order). This is as expected, too, since the function  $\wparen$ does not distinguish the characters $0,1,\dotsc,9$.


\subsection{RQ3: Accelerating Inference Time }
We conducted experiments about  inference time, comparing the original RNNs $R$ and the WFAs $A$ that we extracted from $R$. 
We used the same RNNs $R$ and WFAs $A$ as in~\S{}\ref{subsec:experiment_RQ1}, where the latter are extracted using $\regr{2\text{--}5}$ and $\search{500\text{--}5000}$.
We note that the inference of RNNs utilizes GPUs while that of WFAs is solely done by CPUs.

We observed that 
 the inference time of the extracted WFAs $A$ was about 1,300 times faster than the target RNNs $R$, taking the average over different settings (\ArxivVersion{Appendix~\ref{appendix:inferencetimedetail}}\CameraReadyVersion{Appendix~B.3 of~\cite{DBLP:journals/corr/abs-1904-02931}}). This demonstrates the potential use of the extracted WFAs as a computationally cheaper surrogate for RNNs. We attribute the acceleration to the following: 1) WFAs use only linear computation while RNNs involve nonlinear ones; and 2) overall, extracted WFAs are smaller in size. 
Provided that the accuracy of  extracted WFAs can be  high  (as we observed in \S{}\ref{subsec:experiment_RQ1}), we believe
the replacement of RNNs by WFAs is a viable option  in some application scenarios. 

\section{Conclusions and Future Work}
We proposed a method that extracts a WFA from an RNN, focusing on RNNs that  take a word
$w\in\Sigma^*$ and return a real value.
We used regression to investigate and abstract the internal states of RNNs.  
We experimentally evaluated our method, comparing its performance with a baseline whose equivalence queries are based on simple breadth-first search.

One future work is a detailed comparison with other methods for model compression. 
Another future work is to use machine learning methods to find a counterexample in the equivalence query, such as reinforcement learning~\cite{DBLP:journals/nature/MnihKSRVBGRFOPB15} adversarial attacks~\cite{DBLP:conf/milcom/PapernotMSH16}, and acquisition functions of GPR.
%
Finally, we need a means to optimize parameter $M$ of our method for a
specific problem.  It may also be helpful to extend our method so that the
investigated words can be restricted to a fixed language $L\subset \Sigma^*$; If
$L$ identifies the input space of the training dataset for RNNs, we could avoid
investigating the input space on which the RNNs are not trained, and therefore
we could seek only ``meaningful'' counterexamples even in using large $M$.


\section{Acknowledgments}
Thanks are due to Mahito Sugiyama and the anonymous reviewers of AAAI for a lot of useful comments.
This work is partially supported by
JST ERATO HASUO Metamathematics for Systems Design Project (No.\ JPMJER1603),
JSPS KAKENHI Grant Numbers JP15KT0012, JP18J22498, JP19K20247, JP19K22842, and
JST-Mirai Program Grant Number JPMJMI18BA, Japan.

\bibliography{aaai20}

\begin{thebibliography}{}

\bibitem[\protect\citeauthoryear{Angluin}{1987}]{DBLP:journals/iandc/Angluin87}
Angluin, D.
\newblock 1987.
\newblock Learning regular sets from queries and counterexamples.
\newblock {\em Inf. Comput.} 75(2):87--106.

\bibitem[\protect\citeauthoryear{Ayache, Eyraud, and
  Goudian}{2018}]{DBLP:conf/icgi/AyacheEG18}
Ayache, S.; Eyraud, R.; and Goudian, N.
\newblock 2018.
\newblock Explaining black boxes on sequential data using weighted automata.
\newblock In Unold, O.; Dyrka, W.; and Wieczorek, W., eds., {\em Proc. {ICGI}
  2018}, volume~93 of {\em Proceedings of Machine Learning Research},  81--103.
\newblock {PMLR}.

\bibitem[\protect\citeauthoryear{Baier and Katoen}{2008}]{BaierK08}
Baier, C., and Katoen, J.-P.
\newblock 2008.
\newblock {\em Principles of Model Checking}.
\newblock The MIT Press.

\bibitem[\protect\citeauthoryear{Balle and
  Mohri}{2015}]{DBLP:conf/cai/BalleM15}
Balle, B., and Mohri, M.
\newblock 2015.
\newblock Learning weighted automata.
\newblock In Maletti, A., ed., {\em Proc. {CAI} 2015}, volume 9270 of {\em
  Lecture Notes in Computer Science},  1--21.
\newblock Springer.

\bibitem[\protect\citeauthoryear{Balle, Gourdeau, and
  Panangaden}{2017}]{BalleGP17}
Balle, B.; Gourdeau, P.; and Panangaden, P.
\newblock 2017.
\newblock Bisimulation metrics for weighted automata.
\newblock In Chatzigiannakis, I.; Indyk, P.; Kuhn, F.; and Muscholl, A., eds.,
  {\em Proc. {ICALP} 2017}, volume~80 of {\em LIPIcs},  103:1--103:14.
\newblock Schloss Dagstuhl - Leibniz-Zentrum fuer Informatik.

\bibitem[\protect\citeauthoryear{Bramer}{2013}]{Bramer2013}
Bramer, M.
\newblock 2013.
\newblock {\em Ensemble Classification}.
\newblock London: Springer London.
\newblock  209--220.

\bibitem[\protect\citeauthoryear{Bucila, Caruana, and
  Niculescu{-}Mizil}{2006}]{BucilaCN06}
Bucila, C.; Caruana, R.; and Niculescu{-}Mizil, A.
\newblock 2006.
\newblock Model compression.
\newblock In {\em Proc. KDD 2006},  535--541.

\bibitem[\protect\citeauthoryear{Chaudhuri}{2019}]{DBLP:series/sbcs/Chaudhuri19}
Chaudhuri, A.
\newblock 2019.
\newblock {\em Visual and Text Sentiment Analysis through Hierarchical Deep
  Learning Networks}.
\newblock Springer Briefs in Computer Science. Springer.

\bibitem[\protect\citeauthoryear{Chung \bgroup et al\mbox.\egroup
  }{2014}]{DBLP:journals/corr/ChungGCB14}
Chung, J.; G{\"{u}}l{\c{c}}ehre, {\c{C}}.; Cho, K.; and Bengio, Y.
\newblock 2014.
\newblock Empirical evaluation of gated recurrent neural networks on sequence
  modeling.
\newblock {\em CoRR} abs/1412.3555.

\bibitem[\protect\citeauthoryear{Droste, Kuich, and Vogler}{2009}]{DrosteKV09}
Droste, M.; Kuich, W.; and Vogler, H.
\newblock 2009.
\newblock {\em Handbook of Weighted Automata}.
\newblock Monographs in Theoretical Computer Science. An EATCS Series. Springer
  Berlin Heidelberg.

\bibitem[\protect\citeauthoryear{Du \bgroup et al\mbox.\egroup
  }{2019}]{DBLP:conf/sigsoft/DuXLM0Z19}
Du, X.; Xie, X.; Li, Y.; Ma, L.; Liu, Y.; and Zhao, J.
\newblock 2019.
\newblock Deepstellar: model-based quantitative analysis of stateful deep
  learning systems.
\newblock In Dumas, M.; Pfahl, D.; Apel, S.; and Russo, A., eds., {\em Proc.
  {ESEC/FSE} 2019},  477--487.
\newblock {ACM}.

\bibitem[\protect\citeauthoryear{Gupta \bgroup et al\mbox.\egroup
  }{2015}]{GuptaAGN15}
Gupta, S.; Agrawal, A.; Gopalakrishnan, K.; and Narayanan, P.
\newblock 2015.
\newblock Deep learning with limited numerical precision.
\newblock In {\em Proc. {ICML} 2015},  1737--1746.

\bibitem[\protect\citeauthoryear{Han \bgroup et al\mbox.\egroup
  }{2015}]{HanPTD15}
Han, S.; Pool, J.; Tran, J.; and Dally, W.~J.
\newblock 2015.
\newblock Learning both weights and connections for efficient neural network.
\newblock In {\em Proc. NIPS 2015},  1135--1143.

\bibitem[\protect\citeauthoryear{Hochreiter and
  Schmidhuber}{1997}]{DBLP:journals/neco/HochreiterS97}
Hochreiter, S., and Schmidhuber, J.
\newblock 1997.
\newblock Long short-term memory.
\newblock {\em Neural Computation} 9(8):1735--1780.

\bibitem[\protect\citeauthoryear{Lv \bgroup et al\mbox.\egroup
  }{2018}]{DBLP:journals/access/LvWYL18}
Lv, S.; Wang, J.; Yang, Y.; and Liu, J.
\newblock 2018.
\newblock Intrusion prediction with system-call sequence-to-sequence model.
\newblock {\em {IEEE} Access} 6:71413--71421.

\bibitem[\protect\citeauthoryear{Michalenko \bgroup et al\mbox.\egroup
  }{2019}]{DBLP:conf/iclr/MichalenkoSVBCP19}
Michalenko, J.~J.; Shah, A.; Verma, A.; Baraniuk, R.~G.; Chaudhuri, S.; and
  Patel, A.~B.
\newblock 2019.
\newblock Representing formal languages: {A} comparison between finite automata
  and recurrent neural networks.
\newblock In {\em Proc. {ICLR} 2019}.
\newblock OpenReview.net.

\bibitem[\protect\citeauthoryear{Mnih \bgroup et al\mbox.\egroup
  }{2015}]{DBLP:journals/nature/MnihKSRVBGRFOPB15}
Mnih, V.; Kavukcuoglu, K.; Silver, D.; Rusu, A.~A.; Veness, J.; Bellemare,
  M.~G.; Graves, A.; Riedmiller, M.~A.; Fidjeland, A.; Ostrovski, G.; Petersen,
  S.; Beattie, C.; Sadik, A.; Antonoglou, I.; King, H.; Kumaran, D.; Wierstra,
  D.; Legg, S.; and Hassabis, D.
\newblock 2015.
\newblock Human-level control through deep reinforcement learning.
\newblock {\em Nature} 518(7540):529--533.

\bibitem[\protect\citeauthoryear{Okudono \bgroup et al\mbox.\egroup
  }{2019}]{DBLP:journals/corr/abs-1904-02931}
Okudono, T.; Waga, M.; Sekiyama, T.; and Hasuo, I.
\newblock 2019.
\newblock Weighted automata extraction from recurrent neural networks via
  regression on state spaces.
\newblock {\em CoRR} abs/1904.02931.

\bibitem[\protect\citeauthoryear{Omlin and Giles}{1996}]{OmlinG96}
Omlin, C.~W., and Giles, C.~L.
\newblock 1996.
\newblock Extraction of rules from discrete-time recurrent neural networks.
\newblock {\em Neural Networks} 9(1):41--52.

\bibitem[\protect\citeauthoryear{Papernot \bgroup et al\mbox.\egroup
  }{2016}]{DBLP:conf/milcom/PapernotMSH16}
Papernot, N.; McDaniel, P.~D.; Swami, A.; and Harang, R.~E.
\newblock 2016.
\newblock Crafting adversarial input sequences for recurrent neural networks.
\newblock In Brand, J.; Valenti, M.~C.; Akinpelu, A.; Doshi, B.~T.; and Gorsic,
  B.~L., eds., {\em Proc. {MILCOM} 2016},  49--54.
\newblock {IEEE}.

\bibitem[\protect\citeauthoryear{Rabusseau, Li, and
  Precup}{2019}]{DBLP:conf/aistats/RabusseauLP19}
Rabusseau, G.; Li, T.; and Precup, D.
\newblock 2019.
\newblock Connecting weighted automata and recurrent neural networks through
  spectral learning.
\newblock In Chaudhuri, K., and Sugiyama, M., eds., {\em Proc. {AISTATS} 2019},
  volume~89 of {\em Proceedings of Machine Learning Research},  1630--1639.
\newblock {PMLR}.

\bibitem[\protect\citeauthoryear{Schwartz, Thomson, and
  Smith}{2018}]{schwartz-etal-2018-bridging}
Schwartz, R.; Thomson, S.; and Smith, N.~A.
\newblock 2018.
\newblock Bridging {CNN}s, {RNN}s, and weighted finite-state machines.
\newblock In {\em Proceedings of the 56th Annual Meeting of the Association for
  Computational Linguistics (Volume 1: Long Papers)},  295--305.
\newblock Melbourne, Australia: Association for Computational Linguistics.

\bibitem[\protect\citeauthoryear{Shen, Bao, and
  Huang}{2018}]{shen2018recurrent}
Shen, Z.; Bao, W.; and Huang, D.-S.
\newblock 2018.
\newblock Recurrent neural network for predicting transcription factor binding
  sites.
\newblock {\em Scientific reports} 8(1):15270.

\bibitem[\protect\citeauthoryear{Sutskever, Vinyals, and
  Le}{2014}]{DBLP:conf/nips/SutskeverVL14}
Sutskever, I.; Vinyals, O.; and Le, Q.~V.
\newblock 2014.
\newblock Sequence to sequence learning with neural networks.
\newblock In Ghahramani, Z.; Welling, M.; Cortes, C.; Lawrence, N.~D.; and
  Weinberger, K.~Q., eds., {\em Proc. NIPS 2014},  3104--3112.

\bibitem[\protect\citeauthoryear{Wang and
  Niepert}{2019}]{DBLP:conf/icml/WangN19}
Wang, C., and Niepert, M.
\newblock 2019.
\newblock State-regularized recurrent neural networks.
\newblock In Chaudhuri, K., and Salakhutdinov, R., eds., {\em Proceedings of
  the 36th International Conference on Machine Learning, {ICML} 2019, 9-15 June
  2019, Long Beach, California, {USA}}, volume~97 of {\em Proceedings of
  Machine Learning Research},  6596--6606.
\newblock {PMLR}.

\bibitem[\protect\citeauthoryear{Weiss, Goldberg, and
  Yahav}{2018a}]{DBLP:conf/icml/WeissGY18}
Weiss, G.; Goldberg, Y.; and Yahav, E.
\newblock 2018a.
\newblock Extracting automata from recurrent neural networks using queries and
  counterexamples.
\newblock In Dy, J.~G., and Krause, A., eds., {\em Proc. {ICML} 2018},
  volume~80 of {\em {JMLR} Workshop and Conference Proceedings},  5244--5253.
\newblock JMLR.org.

\bibitem[\protect\citeauthoryear{Weiss, Goldberg, and
  Yahav}{2018b}]{DBLP:conf/acl/WeissGY18}
Weiss, G.; Goldberg, Y.; and Yahav, E.
\newblock 2018b.
\newblock On the practical computational power of finite precision rnns for
  language recognition.
\newblock In Gurevych, I., and Miyao, Y., eds., {\em Proc. ACL 2018, Volume 2:
  Short Papers},  740--745.
\newblock Association for Computational Linguistics.

\bibitem[\protect\citeauthoryear{Xiao \bgroup et al\mbox.\egroup
  }{2019}]{DBLP:journals/mta/XiaoZMHS19}
Xiao, X.; Zhang, S.; Mercaldo, F.; Hu, G.; and Sangaiah, A.~K.
\newblock 2019.
\newblock Android malware detection based on system call sequences and {LSTM}.
\newblock {\em Multimedia Tools Appl.} 78(4):3979--3999.

\bibitem[\protect\citeauthoryear{Yarowsky}{1995}]{Yarowsky95}
Yarowsky, D.
\newblock 1995.
\newblock Unsupervised word sense disambiguation rivaling supervised methods.
\newblock In {\em Proc. ACL 1995},  189--196.

\bibitem[\protect\citeauthoryear{Zweig \bgroup et al\mbox.\egroup
  }{2017}]{DBLP:conf/icassp/ZweigYDS17}
Zweig, G.; Yu, C.; Droppo, J.; and Stolcke, A.
\newblock 2017.
\newblock Advances in all-neural speech recognition.
\newblock In {\em Proc. {ICASSP} 2017},  4805--4809.
\newblock {IEEE}.

\end{thebibliography}
\bibliographystyle{aaai}

\ifdefined \VersionCameraReady
\else
\appendix

\auxproof{\input{extra_preliminaries}}

\auxproof{\section{NIPS preliminaries}

The formalization of deterministic finite automata (DFAs) and
weighted finite automata (WFAs) is recalled, mainly  to fix
notations. See e.g.,~\cite{DrosteKV09} for details.
In what follows, \emph{matrices} and \emph{vectors} refer to those over the set of real numbers $\R$.



 A \emph{weighted finite automaton} (WFA) over $\Sigma$  is
 a quadruple 
\begin{math}
 A=\bigl(Q_A, \alpha_A, \beta_A, (A_\sigma)_{\sigma\in \Sigma}\bigr)
\end{math}.
Here $Q_A$ is a finite set of \emph{states}; $\alpha_A, \beta_A$ are row vectors of size $|Q_{A}|$ called the \emph{initial} and \emph{final} vectors; and $A_\sigma$  is a
 \emph{transition} matrix of $\sigma$, given for each $\sigma\in\Sigma$. The matrix size of $A_\sigma$ is $|Q_{A}|\times|Q_{A}|$.

A \emph{configuration} of $A$ is a row vector $x\in \R^{Q_{A}}$. 
Given a word $w=\sigma_{1}\sigma_{2}\dotsc \sigma_{n}\in \Sigma^{*}$ (where $\sigma_{i}\in \Sigma$), the \emph{configuration} of $A$ at $w$ is defined by 
\begin{math}
\textstyle \delta_A(w) = \alpha_A^\top \cdot \bigl(\prod_{i=1}^{n} A_{\sigma_i}\bigr)
\end{math}.
The \emph{weight} of $w$ in $A$ is defined by
\begin{math}
\textstyle f_A(w) = \alpha_A^\top \cdot\bigl(\prod_{i=1}^n A_{\sigma_i}\bigr)\cdot \beta_A\,\in\R
\end{math},
multiplying the final vector to the configuration at $w$.

A WFA $A$ is \emph{probabilistic} if all the elements of $\alpha_A, \beta_A, (A_\sigma)_{\sigma \in \Sigma}$ are in $[0, 1]$, $\sum_{q \in Q_A} (\alpha_A)_q = 1$ and $\sum_{c\in Q_A} (A_\sigma)_{r, c}=1$ for all $r\in Q_A$.
\TS{We may note why the final vector is not normalized.}

A WFA $A$ is \emph{minimal} if there is no WFA $A'=(Q_{A'}, \alpha_{A'}, \beta_{A'}, (A'_\sigma)_{\sigma\in \Sigma})$ such that $\abs{Q_{A'}} < \abs{Q_A}$.
\TS{This definition seems not to be finished.}

\subsubsection{Recurrent Neural Networks (RNNs)}
 A (real-valued) \emph{recurrent neural network} (RNN) of \emph{dimension} $d\in\N$ is a triple
\begin{math}
 R=(\alpha_R, \beta_R, g_R)
\end{math},
where $\alpha_R\in \R^{d}$ is an \emph{initial state}, 
$\beta_R\colon \R^d\to \R$ is an \emph{output function}, and $g_R\colon \R^d\times \Sigma \to \R^d$ is called a \emph{transition function}. The set $\R^{d}$ is called a \emph{state space}. 
The transition function $g_R$  naturally extends to words: $g^{*}_R\colon \R^d\times \Sigma^* \to \R^d$, defined inductively by
$ g^{*}_R(x, \varepsilon) = x$ and
$
  g^{*}_R(x, w\sigma) = g_R\bigl(g^{*}_R(x, w), \sigma\bigr)
$, where $w\in \Sigma^{*}$ and $\sigma\in \Sigma$.
The \emph{configuration} $\delta_{R}(w)$ of the RNN $R$ at a word $w$ is defined by 
$\delta_R(w) = g^{*}_{R}(\alpha_R, w)$. The \emph{output} $f_{R}(w)\in \R$, of $R$ for the input $w$, is defined by $f_{R}(w)=\beta_{R}\bigl(\delta_R(w)\bigr)$.

\subsubsection{\lstar~Algorithm for WFA Learning}
\label{subsec:balle}
Angluin's \lstar~algorithm is an algorithm to learn a DFA \cite{DBLP:journals/iandc/Angluin87}, utilizing \emph{membership query} and \emph{equivalence query}.
Given a word, \emph{membership query} answers if it is accepted by the target DFA.  Given a DFA, \emph{equivalence query} answers if it is equivalent to the target DFA, and \taro{may?} returns a counterexample if they are not equivalent.
The classic \lstar~algorithm has seen a \emph{weighted} extension~\cite{DBLP:conf/cai/BalleM15}: it learns a WFA $B$, again via a series of membership and equivalence queries.
In the literature~\cite{DBLP:conf/cai/BalleM15}, an observation table $T$ is presented as a so-called \emph{Hankel} matrix. This opens the way to further extensions of the method, such as an approximate learning algorithm via the singular-value decomposition (SVD).

\auxproof{Equivalence of $\mathbb{Q}$-weighted automata seems to be decidable. See~\cite{KieferMOWW13}.}
}

\section{Detail of Our WFA Extraction}

\subsection{On Rank Tolerance}\label{appendix:rankTolerance}
Construction step calculates a minimal WFA that is compatible with observed data $(w_1, f_R(w_1)), \dots, (w_n, f_R(w_n)) \in \Sigma^*\times \R$.  It relies on rank calculation, and the calculation is done by computing the SVD of the matrix and counting the number of non-zero singular values.  The threshold to check whether the singular value is zero or not is called \emph{rank tolerance}.  A small rank tolerance results in accurate learning basically but can cause overfitting for short words and huge error for long words.  A large rank tolerance results in rough learning but prevents such overfitting.
To balance the rank tolerance, we start from a big initial rank tolerance $\tau$, and if it is too big then we decay it by multiplying $r (0<r<1)$.
We know that the rank tolerance is too big if the equivalence query returns the same counterexample twice because it means the counterexample was ignored.
Overall, we obtain the WFA Extraction procedure (Algorithm~\ref{alg:zentai}).

\begin{algorithm}[bpt]
   \caption{WFA Extraction}
   \label{alg:zentai}
\begin{algorithmic}[1]
 \State {\bfseries Input:} Target RNN $R$, initial rank tolerance $\tau$, 
 decay rate of rank tolerance $r(0<r<1)$
 \State{$\mathtt{result} \gets \mathtt{None}$}
 \State{$\mathtt{result}' \gets \mathtt{None}$}
 \State{$\mathtt{observed} \gets \set{(\varepsilon, f_R(\varepsilon))}$} 
 \State{Construct a WFA $A$ by $\mathtt{observed}$}
 \Loop
 \State{$\mathtt{result}' \gets (\text{result of equivalence query for $A$})$ }
 \State{\textbf{insert} $(result', f_R(result'))$ \textbf{to} $\mathtt{observed}$}
 \If{$\mathtt{result} = \mathtt{result}'$}\par
 \Comment{The latest counterexample did not work.}
 \State{$\tau \gets r\tau$} \label{line:zentai-decay}
 \Comment{Decrease the rank tolerance $\tau$}
 \Else
 \State{Update $A$ by $(\mathtt{result}', f_R(\mathtt{result}'))$ with $\tau$}\par
 \Comment{$\tau$ is the current rank tolerance.}
 \EndIf
 \If{$\mathtt{result} = \text{ $\mathtt{Equivalent}$}$}
 \Break
 \EndIf
 \State{$\mathtt{result} \gets \mathtt{result}'$}
 \EndLoop
 \Return{$A$}
\end{algorithmic}
\end{algorithm}

\section{Detail of the Experiments}

\subsection{On Training Data Generation for $\wparen$}
\label{appendix:wparenDataGen}
We made the training data for $\wparen$ in this manner.
\begin{enumerate}
  \item We make 5000 words of random balanced parentheses made only of $\set{(, )}$.
  There is a one-to-one correspondence between words of balanced parentheses of length $2n$ and paths from the bottom-left to the top-right in the grid of size $n\times n$ whose bottom-right half is removed, so we can obtain such random words by generating the paths randomly and converting them into the words.  For example, ``(())'' or ``(()())'' can be made.
  \item We insert random characters in $\set{0, 1, \dots, 9}$ into the words generated in Step 1.
  This generates 5000 words of random balanced words made of $\set{(, ), 0, 1, \dots, 9}$.  For example, ``(0(1))'' or ``((12340)())'' can be made.
  \item We run the same procedure as Step 1 and obtain 5000 words of random balanced parentheses.
  \item We mutate the words in Step 3 and make them into 5000 random unbalanced parentheses made only of $\set{(, )}$. The mutation rules are as follows: 1) duplicate a random character; 2) delete a random character; and 3) exchange a random pair of adjacent characters. 
 These rules are repeatedly applied---each time throwing a fair coin---until we get the head of the coin.
Note that the mutation can make a balanced word into another balanced word.  For example, ``(()'', ``(((('', or ``()'' can be made (only the last one is balanced).
  \item We insert random characters in $\set{0, 1, \dots, 9}$ into the words generated in Step 4.
  This generates 5000 words of random unbalanced words made of $\set{(, ), 0, 1, \dots, 9}$. 
  \item We combine the result of Step 2 and 5 and get 10000 words.  Almost the half of the words are balanced and the other half are unbalanced.  We pick 9000 random words from the words and use them as the training data; the remaining 1000 are used as the test data.
\end{enumerate}

\subsection{Detailed WFAs Extracted from $\wparen$}
\label{appendix:wparenWFADetail}
\subsubsection{The WFA Extracted by $\regr{5}$}
\begin{figure*}[tb]
 \centering
  \includegraphics[width=15.0cm]{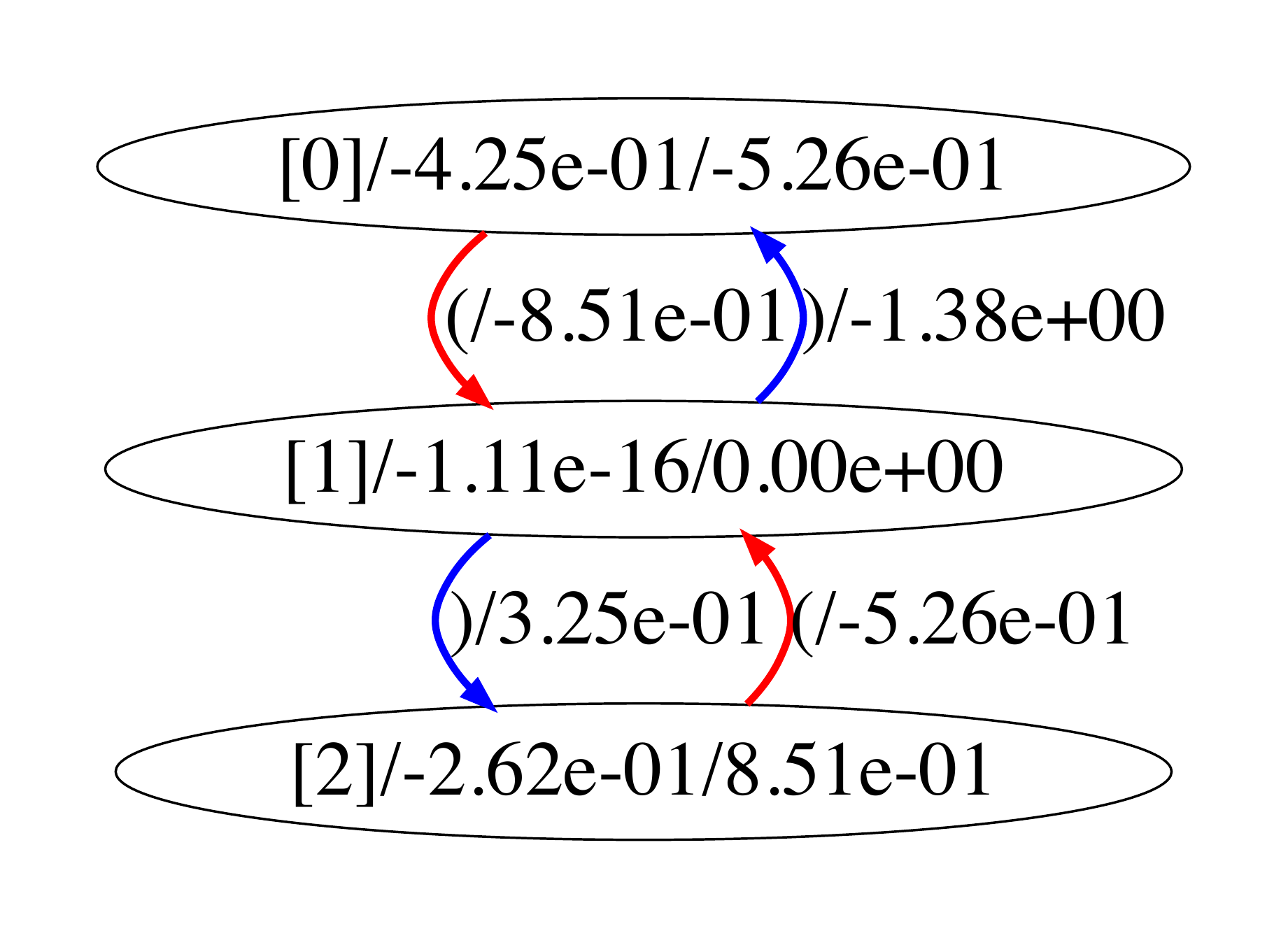}
 \caption{The extracted WFAs by $\regr{5}$}
 \label{fig:WFA_regr5_raw}
\end{figure*}
\begin{figure*}[tbp]
 \centering
 \begin{align*}
 \WFAInitState &= 
 \begin{pmatrix}
  -4.24538470e-01\\ -1.10824765e-16\\ -2.62446661e-01
 \end{pmatrix}\\
 \WFAAccState &= 
 \begin{pmatrix}
  -0.52582891 \\ 0         \\ 0.85059036
 \end{pmatrix}
 \end{align*} 
 \caption{The initial and accepting vector of the extracted WFAs by $\regr{5}$}
 \label{fig:WFA_regr5_initial_accepting_vector}
\end{figure*}
\begin{figure*}[tbp]
\centering
 \begin{align*}
\WFATransition[(] & =
\begin{pmatrix}
 0.00000000e+00 & -8.50590360e-01 & 0.00000000e+00\\
 0.00000000e+00 & -8.59785403e-17 & 0.00000000e+00\\
 0.00000000e+00 & -5.25828907e-01 & 0.00000000e+00
\end{pmatrix}\\
\WFATransition[)] & =
\begin{pmatrix}
 0.00000000e+00 & 0.00000000e+00 & 0.00000000e+00\\
 -1.37593037e+00 & 0.00000000e+00 & 3.25063688e-01\\
\end{pmatrix}\\
\WFATransition[0] & =
\begin{pmatrix}
9.97100655e-01 & 0.00000000e+00 & -6.69023024e-04\\
2.91425732e-19 & 9.99289799e-01 & 1.63393525e-16\\
1.78230607e-03 & -3.85026199e-16 & 9.99284687e-01\\
\end{pmatrix}\\
\WFATransition[1] & =
\begin{pmatrix}
9.97754767e-01 & 0.00000000e+00 & -1.10479143e-03\\
1.51485831e-19 & 1.00062565e+00 & 1.63529229e-16\\
9.26459427e-04 & -3.85540902e-16 & 1.00011463e+00\\
\end{pmatrix}\\
\WFATransition[2] & =
\begin{pmatrix}
9.97342696e-01 & 0.00000000e+00 & -7.41831462e-04\\
2.47685043e-19 & 1.00090522e+00 & 1.63425319e-16\\
1.51479608e-03 & -3.85648619e-16 & 9.99479132e-01\\
\end{pmatrix}\\
\WFATransition[3] & =
\begin{pmatrix}
9.97152196e-01 & 0.00000000e+00 & -5.48790930e-04\\
2.62410313e-19 & 1.00054480e+00 & 1.63352215e-16\\
1.60485312e-03 & -3.85509751e-16 & 9.99032039e-01\\
\end{pmatrix}\\
\WFATransition[4] & =
\begin{pmatrix}
9.97528034e-01 & 0.00000000e+00 & -9.46331754e-04\\
2.49455242e-19 & 1.00138911e+00 & 1.63510808e-16\\
1.52562229e-03 & -3.85835064e-16 & 1.00000197e+00\\
\end{pmatrix}\\
\WFATransition[5] & =
\begin{pmatrix}
9.97514811e-01 & 0.00000000e+00 & -1.16262080e-03\\
1.67099898e-19 & 9.99657498e-01 & 1.63514942e-16\\
1.02195218e-03 & -3.85167873e-16 & 1.00002725e+00\\
\end{pmatrix}\\
\WFATransition[6] & =
\begin{pmatrix}
9.98021793e-01 & 0.00000000e+00 & -1.24638991e-03\\
1.41062546e-19 & 9.99932885e-01 & 1.63603900e-16\\
8.62712533e-04 & -3.85273980e-16 & 1.00057130e+00\\
\end{pmatrix}\\
\WFATransition[7] & =
\begin{pmatrix}
9.97351931e-01 & 0.00000000e+00 & -1.02957965e-03\\
2.34439773e-19 & 1.00063437e+00 & 1.63494750e-16\\
1.43379045e-03 & -3.85544261e-16 & 9.99903757e-01\\
\end{pmatrix}\\
\WFATransition[8] & =
\begin{pmatrix}
9.97459514e-01 & 0.00000000e+00 & -7.96005125e-04\\
2.27400184e-19 & 1.00190830e+00 & 1.63446209e-16\\
1.39073761e-03 & -3.86035107e-16 & 9.99606890e-01\\
\end{pmatrix}\\
\WFATransition[9] & =
\begin{pmatrix}
9.97267117e-01 & 0.00000000e+00 & -5.40868994e-04\\
3.80598316e-19 & 9.99770233e-01 & 1.63441973e-16\\
2.32766916e-03 & -3.85211310e-16 & 9.99580985e-01\\
\end{pmatrix}\\
\end{align*}
 \caption{The transition matrices of the extracted WFAs by $\regr{5}$}
 \label{fig:WFA_regr5_transition_matrices}
\end{figure*}
Fig.~\ref{fig:WFA_regr5_raw} illustrates the WFA extracted from the RNN trained by $\wparen$ by $\regr{5}$.
The initial and final vectors, and the transition matrices are in Fig.~\ref{fig:WFA_regr5_initial_accepting_vector} and~\ref{fig:WFA_regr5_transition_matrices}.

\subsubsection{The WFA Extracted by $\regr{15}$}
\begin{figure*}[tb]
 \centering
  \includegraphics[width=15.0cm]{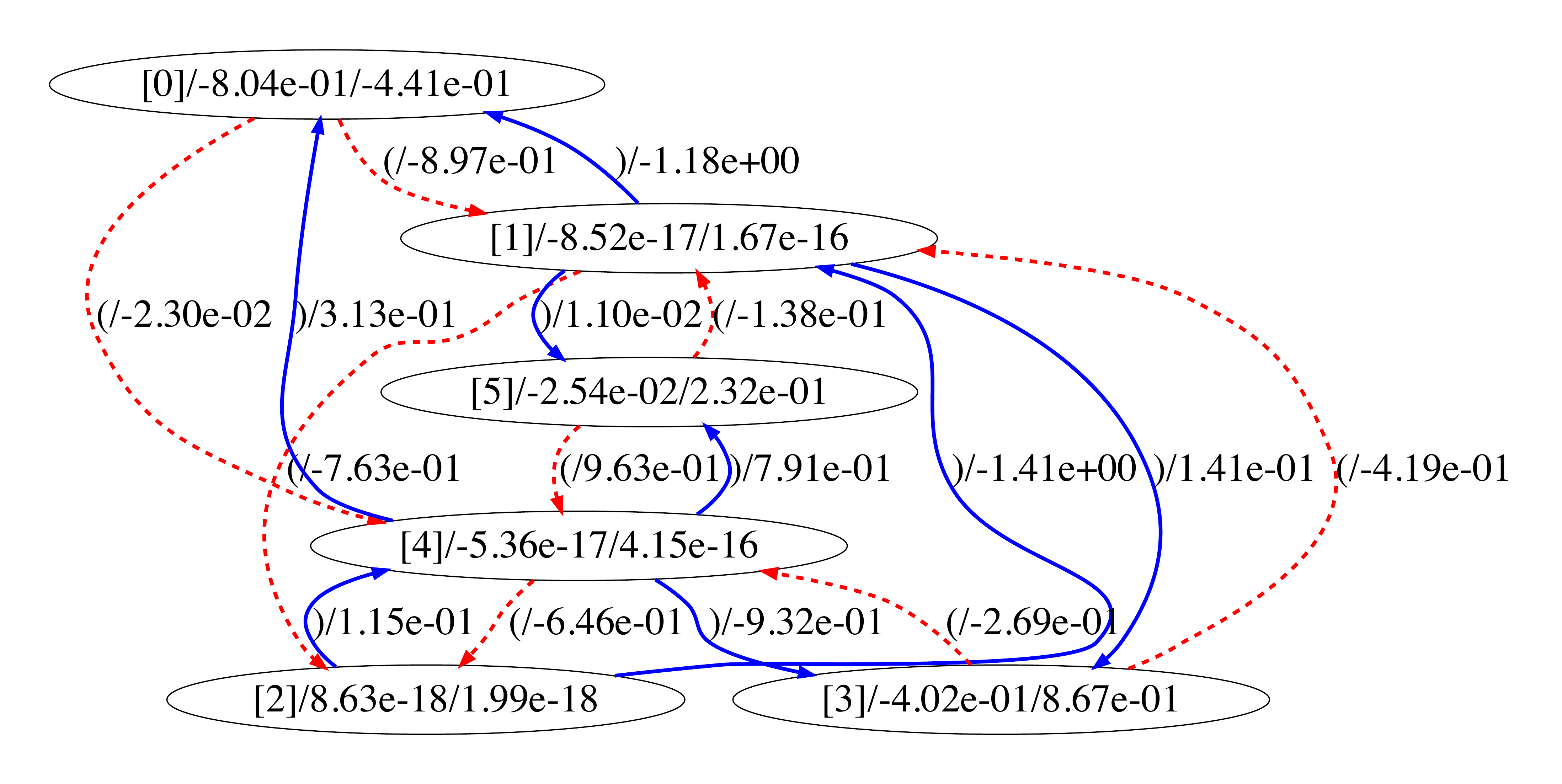}
 \caption{The extracted WFAs by $\regr{15}$}
 \label{fig:WFA_regr15_raw}
\end{figure*}
\begin{figure*}[tbp]
 \centering
 \begin{align*}
  \WFAInitState &= 
  \begin{pmatrix}
   -8.03989494e-01 \\ 
   -8.52463864e-17 \\ 8.62805331e-18 \\ -4.02164050e-01 \\ -5.35508802e-17 \\ -2.54365543e-02
  \end{pmatrix}\\
  \WFAAccState &=
  \begin{pmatrix}
   -4.41053365e-01 \\ 1.66533454e-16 \\ 1.99459802e-18 \\ 8.67087989e-01 \\ 4.14765037e-16 \\ 2.31582271e-01
  \end{pmatrix}
 \end{align*}
 \caption{The initial and accepting vector of the extracted WFAs by $\regr{15}$}
 \label{fig:WFA_regr15_initial_accepting_vector}
\end{figure*}
\begin{figure*}[tbp]
\centering
\scriptsize
\begin{align*}
\WFATransition[(] & =
\begin{pmatrix}
-8.13455245e-17 & -8.97187248e-01 & -9.14353577e-17 & 6.05261178e-17 & -2.29558342e-02 & 2.63293240e-16\\
4.66228882e-33 & 4.35160982e-16 & -7.63485724e-01 & -8.48075160e-17 & 2.11401164e-16 & -2.35492323e-16\\
-5.15551962e-33 & -4.68274918e-17 & 1.26662796e-16 & 1.62130062e-32 & 1.23274983e-17 & 4.28769662e-32\\
-2.06614561e-17 & -4.19377765e-01 & 1.32499255e-16 & 4.76754407e-17 & -2.68850720e-01 & 3.12643472e-16\\
-3.44414230e-32 & 1.23963211e-16 & -6.45824705e-01 & -7.17378037e-17 & 4.14160693e-16 & -1.99200529e-16\\
-7.75638689e-17 & -1.38482243e-01 & 1.60067044e-15 & -6.32326215e-17 & 9.62908262e-01 & -6.69148936e-16
\end{pmatrix}\\
\WFATransition[)] & =
\begin{pmatrix}
-5.54148872e-17 & -1.97207974e-20 & 2.03671709e-33 & 7.46642668e-18 & 1.61218726e-21 & -2.61486321e-19\\
-1.18211303e+00 & 2.82084078e-16 & 2.75198255e-17 & 1.40589069e-01 & -3.46762135e-17 & 1.09968804e-02\\
1.29599806e-16 & -1.41195049e+00 & -5.84247592e-17 & 3.14911192e-18 & 1.15427817e-01 & 3.43913931e-16\\
2.33225409e-16 & -7.98892997e-17 & 4.54190736e-32 & 3.57233072e-17 & 6.53099913e-18 & -5.84636207e-17\\
3.13336842e-01 & -1.07422125e-15 & -7.71191144e-16 & -9.32439297e-01 & 4.01725110e-16 & 7.91163583e-01\\
1.01495616e-15 & -8.56767927e-16 & 6.25086145e-31 & 6.81033112e-16 & 7.00413022e-17 & -7.20640086e-16
\end{pmatrix}\\
\WFATransition[0] & =
\begin{pmatrix}
9.99296632e-01 & -5.59800079e-17 & 4.40354106e-17 & 3.98075885e-04 & 6.13728232e-17 & -2.01408442e-04\\
1.15541074e-17 & 1.00082887e+00 & -1.21731926e-17 & -4.17682688e-18 & 7.78834199e-04 & -2.67091700e-16\\
2.98176861e-19 & -7.22564866e-16 & 1.00029933e+00 & 2.09488298e-16 & -2.31000082e-16 & 3.49208348e-16\\
-1.56179092e-03 & 2.15918979e-16 & 6.43243081e-16 & 9.98434636e-01 & 2.04617242e-16 & -1.60951935e-03\\
1.17519561e-16 & 1.17507940e-03 & 6.82685692e-16 & -5.63812348e-17 & 1.00486186e+00 & -4.86957236e-16\\
1.32703493e-02 & -1.04785834e-15 & 3.03641615e-16 & 9.09412823e-05 & -1.50126714e-16 & 1.01293307e+00
\end{pmatrix}\\
\WFATransition[1] & =
\begin{pmatrix}
9.99407998e-01 & -5.60117474e-17 & 4.39625414e-17 & 3.24682067e-04 & 6.14442452e-17 & -8.59144355e-05\\
1.15304960e-17 & 1.00102065e+00 & -1.22272583e-17 & -4.15177985e-18 & 6.91564023e-04 & -2.67083620e-16\\
1.30843906e-19 & -7.22766725e-16 & 1.00097799e+00 & 2.09749493e-16 & -2.31366661e-16 & 3.49447287e-16\\
-1.64676070e-03 & 2.15504979e-16 & 6.43087712e-16 & 9.98109329e-01 & 2.04786438e-16 & -1.61269869e-03\\
1.17108223e-16 & 4.56507250e-03 & 6.82503884e-16 & -5.50183867e-17 & 1.00135870e+00 & -4.85110292e-16\\
9.32707190e-03 & -1.04832190e-15 & 3.06831309e-16 & 5.49548865e-03 & -1.45088795e-16 & 1.01367566e+00
\end{pmatrix}\\
\WFATransition[2] & =
\begin{pmatrix}
9.99293546e-01 & -5.59120244e-17 & 4.40553625e-17 & 5.06854606e-04 & 6.14996619e-17 & -5.55034262e-05\\
1.15903828e-17 & 1.00119303e+00 & -1.22194261e-17 & -4.19667887e-18 & 4.98702032e-04 & -2.66997709e-16\\
2.26523367e-19 & -7.22509804e-16 & 1.00045605e+00 & 2.09575372e-16 & -2.31089602e-16 & 3.49280410e-16\\
-1.60081145e-03 & 2.15324516e-16 & 6.42962343e-16 & 9.97884313e-01 & 2.04570862e-16 & -1.72690027e-03\\
1.17588225e-16 & 5.06284191e-03 & 6.82602341e-16 & -5.57205463e-17 & 1.00295379e+00 & -4.86382094e-16\\
1.15881399e-02 & -1.05019147e-15 & 3.05204965e-16 & 3.16454535e-03 & -1.46713771e-16 & 1.01380881e+00
\end{pmatrix}\\
\WFATransition[3] & =
\begin{pmatrix}
9.99281268e-01 & -5.59573383e-17 & 4.40180842e-17 & 3.78557978e-04 & 6.13506059e-17 & -1.84068466e-04\\
1.16388481e-17 & 1.00142057e+00 & -1.23070729e-17 & -4.17728883e-18 & 1.77934070e-04 & -2.66818663e-16\\
2.71166106e-19 & -7.23094524e-16 & 1.00097409e+00 & 2.09527252e-16 & -2.31028483e-16 & 3.49375582e-16\\
-1.70720401e-03 & 2.15879244e-16 & 6.43231698e-16 & 9.98412235e-01 & 2.04758028e-16 & -1.50939736e-03\\
1.17289083e-16 & 7.79522548e-05 & 6.83082252e-16 & -5.65721181e-17 & 1.00553729e+00 & -4.87459412e-16\\
1.29556596e-02 & -1.04791865e-15 & 3.03922900e-16 & -7.47388994e-04 & -1.51257632e-16 & 1.01167398e+00
\end{pmatrix}\\
\WFATransition[4] & =
\begin{pmatrix}
9.99272730e-01 & -5.58843634e-17 & 4.40804524e-17 & 5.68760210e-04 & 6.15529753e-17 & -2.10161070e-05\\
1.15820152e-17 & 1.00117000e+00 & -1.21945297e-17 & -4.20586579e-18 & 6.86122171e-04 & -2.67119334e-16\\
1.91336929e-19 & -7.22479004e-16 & 1.00061898e+00 & 2.09693820e-16 & -2.31351687e-16 & 3.49382751e-16\\
-1.30996389e-03 & 2.15029536e-16 & 6.42794941e-16 & 9.97582788e-01 & 2.04723407e-16 & -1.68588444e-03\\
1.17597980e-16 & 7.06022786e-03 & 6.82149076e-16 & -5.48032346e-17 & 9.99733713e-01 & -4.84262887e-16\\
1.00042919e-02 & -1.05033502e-15 & 3.06771331e-16 & 6.39114790e-03 & -1.42355472e-16 & 1.01500077e+00
\end{pmatrix}
\end{align*}
 \caption{The transition matrices of the extracted WFAs by $\regr{15}$ (for $\sigma \in \{\texttt{(},\texttt{)},1,2,3,4\}$)}
 \label{fig:WFA_regr15_transition_matrices_no1}
\end{figure*}
\begin{figure*}[tbp]
\centering
\scriptsize
\begin{align*}
\WFATransition[5] & =
\begin{pmatrix}
9.99463469e-01 & -5.61082820e-17 & 4.38814586e-17 & 1.22067377e-04 & 6.12978077e-17 & -2.04093919e-04\\
1.15318176e-17 & 1.00100869e+00 & -1.21912117e-17 & -4.11424946e-18 & 6.36064247e-04 & -2.67033033e-16\\
2.02699532e-19 & -7.22607534e-16 & 1.00029543e+00 & 2.09623345e-16 & -2.30961854e-16 & 3.49224654e-16\\
-1.93123658e-03 & 2.16064799e-16 & 6.43419914e-16 & 9.98755732e-01 & 2.04693783e-16 & -1.62514201e-03\\
1.17159685e-16 & 1.05557406e-03 & 6.82702060e-16 & -5.57875540e-17 & 1.00498703e+00 & -4.86899833e-16\\
1.18051058e-02 & -1.04665878e-15 & 3.04945578e-16 & 2.66653121e-03 & -1.49108601e-16 & 1.01340563e+00
\end{pmatrix}\\
\WFATransition[6] & =
\begin{pmatrix}
9.99270606e-01 & -5.58880977e-17 & 4.41098581e-17 & 5.54905278e-04 & 6.14469389e-17 & -1.46379686e-04\\
1.15941865e-17 & 1.00118061e+00 & -1.22843592e-17 & -4.16283361e-18 & 5.63901004e-04 & -2.67024145e-16\\
7.64012831e-20 & -7.23401942e-16 & 1.00155801e+00 & 2.09960987e-16 & -2.31395681e-16 & 3.49616662e-16\\
-6.63387748e-04 & 2.15542630e-16 & 6.42800551e-16 & 9.97574678e-01 & 2.04705718e-16 & -1.44911555e-03\\
1.15518342e-16 & -1.02797853e-03 & 6.82862561e-16 & -5.35407556e-17 & 1.00336556e+00 & -4.85213752e-16\\
5.57258016e-03 & -1.04098887e-15 & 3.10119657e-16 & 1.04511511e-02 & -1.46075233e-16 & 1.01303930e+00
\end{pmatrix}\\
\WFATransition[7] & =
\begin{pmatrix}
9.99237402e-01 & -5.58952374e-17 & 4.40907525e-17 & 5.48695123e-04 & 6.14799665e-17 & -9.97701660e-05\\
1.15849561e-17 & 1.00108512e+00 & -1.21853831e-17 & -4.18844928e-18 & 6.02383979e-04 & -2.67037089e-16\\
2.12971526e-19 & -7.22355438e-16 & 1.00017887e+00 & 2.09645978e-16 & -2.31039310e-16 & 3.49230916e-16\\
-1.36321195e-03 & 2.15240106e-16 & 6.42861686e-16 & 9.97673235e-01 & 2.04443137e-16 & -1.82208222e-03\\
1.17346570e-16 & 4.88068735e-03 & 6.82433269e-16 & -5.49430977e-17 & 1.00282769e+00 & -4.85969900e-16\\
1.06710808e-02 & -1.04844637e-15 & 3.06438734e-16 & 6.19596029e-03 & -1.45096416e-16 & 1.01493448e+00
\end{pmatrix}\\
\WFATransition[8] & =
\begin{pmatrix}
9.99485693e-01 & -5.60463491e-17 & 4.38945978e-17 & 2.23945615e-04 & 6.14354582e-17 & -4.53253526e-05\\
1.15874726e-17 & 1.00140298e+00 & -1.21941168e-17 & -4.15369161e-18 & 3.55804959e-04 & -2.66949792e-16\\
2.41905849e-19 & -7.22115784e-16 & 9.99859058e-01 & 2.09479615e-16 & -2.30973028e-16 & 3.49093881e-16\\
-2.07582058e-03 & 2.15463744e-16 & 6.43310294e-16 & 9.98594124e-01 & 2.04913786e-16 & -1.67941050e-03\\
1.18327579e-16 & 7.51634097e-03 & 6.82147238e-16 & -5.64154854e-17 & 1.00141082e+00 & -4.85963575e-16\\
1.31311788e-02 & -1.05385406e-15 & 3.03514970e-16 & 6.98588331e-04 & -1.45893667e-16 & 1.01363345e+00
\end{pmatrix}\\
\WFATransition[9] & =
\begin{pmatrix}
9.99315111e-01 & -5.59700858e-17 & 4.39955085e-17 & 3.37854024e-04 & 6.13374084e-17 & -1.83554695e-04\\
1.16759138e-17 & 1.00149706e+00 & -1.22979062e-17 & -4.16208693e-18 & 1.23487542e-04 & -2.66766491e-16\\
3.37885984e-19 & -7.23262350e-16 & 1.00075052e+00 & 2.09528596e-16 & -2.30850079e-16 & 3.49372665e-16\\
-1.33702792e-03 & 2.16271556e-16 & 6.43198414e-16 & 9.98592080e-01 & 2.05060444e-16 & -1.03477715e-03\\
1.16912096e-16 & -4.78824593e-03 & 6.83072626e-16 & -5.63467680e-17 & 1.00742399e+00 & -4.87380576e-16\\
1.37061163e-02 & -1.04360630e-15 & 3.03799727e-16 & -5.37634224e-04 & -1.52428625e-16 & 1.01215221e+00
\end{pmatrix}
\end{align*}
 \caption{The transition matrices of the extracted WFAs by $\regr{15}$ (for $\sigma \in \{5,6,7,8,9\}$)}
 \label{fig:WFA_regr15_transition_matrices_no2}
\end{figure*}
Fig.~\ref{fig:WFA_regr15_raw} illustrates the WFA extracted from the RNN trained by $\wparen$ by $\regr{15}$.
The initial and final vectors, and the transition matrices are in Fig.~\ref{fig:WFA_regr15_initial_accepting_vector}, \ref{fig:WFA_regr15_transition_matrices_no1}, and~\ref{fig:WFA_regr15_transition_matrices_no2}.

\subsection{Inference Time of the Target RNNs and the Extracted WFAs}
\label{appendix:inferencetimedetail}
On average, the inference time of the target RNNs was 29.97519233 milliseconds, 
while that of the extracted WFAs was 0.023052549 milliseconds.
Therefore, on average, the inference of the extracted WFAs was
about 1300.298397 times faster than that of the target RNNs.
\fi

\end{document}